\documentclass[11pt]{article}
\usepackage[utf8]{inputenc}
	\newcommand{\blind}{0}
	
    \makeatletter
    \renewcommand\section{\@startsection {section}{1}{\z@}%
                                       {-3.5ex \@plus -1ex \@minus -.2ex}%
                                       {2.3ex \@plus.2ex}%
                                       {\normalfont\fontfamily{phv}\fontsize{16}{19}\bfseries}}
    \renewcommand\subsection{\@startsection{subsection}{2}{\z@}%
                                         {-3.25ex\@plus -1ex \@minus -.2ex}%
                                         {1.5ex \@plus .2ex}%
                                         {\normalfont\fontfamily{phv}\fontsize{14}{17}\bfseries}}
    \renewcommand\subsubsection{\@startsection{subsubsection}{3}{\z@}%
                                        {-3.25ex\@plus -1ex \@minus -.2ex}%
                                         {1.5ex \@plus .2ex}%
                                         {\normalfont\normalsize\fontfamily{phv}\fontsize{14}{17}\selectfont}}
    \makeatother
	
	\usepackage{amsmath}
	\usepackage{graphicx}
	\usepackage{enumerate}
    \usepackage{soul}

    \usepackage{booktabs}
    \usepackage{multirow}
    \usepackage{siunitx}
    \usepackage{amssymb}
    \usepackage{bbm}
	\usepackage{natbib} 
    \usepackage[utf8]{inputenc}  
    \usepackage[T1]{fontenc}     
    \usepackage{textcomp}        
    \usepackage{graphicx}
    \usepackage{subcaption}
    \usepackage{caption}

	\usepackage{url} 
    \usepackage{xcolor}

\usepackage{hyperref} 
\hypersetup{
	colorlinks=true,       
	linkcolor=blue,        
	citecolor=blue,        
	filecolor=magenta,     
	urlcolor=blue
}
	
\begin{document}
		
		\def\spacingset#1{\renewcommand{\baselinestretch}%
			{#1}\small\normalsize} \spacingset{1}
		
		\if0\blind
		{   
			\title{ \vspace{-0.8in} \bf{Diffusion-corrected Autoregressive Fourier Neural Operator for Droplet Evolution Prediction}}
			\author{Jinghao Cao$^a$, Minsung Kang$^b$, Hongyue Sun$^b$, Chi Zhou$^c$, Jihoon Chung$^d$, \\ Xubo Yue$^e$, Sanchoy Das$^a$ and Bo Shen$^a$ \\
			\small $^a$New Jersey Institute of Technology, USA;
             $^b$University of Georgia, USA; \\
            \small $^c$University at Buffalo, USA; 
              $^d$Hanyang University, South Korea; 
               $^e$Northeastern University, USA
             }
			\date{\vspace{-0.8in}}
			\maketitle
		} \fi
		
		\if1\blind
		{

            \title{\bf \emph{Diffusion-corrected Autoregressive Fourier Neural Operator for Droplet Evolution Prediction} \LaTeX \ Template}
			\author{Author information is purposely removed for double-blind review}
			
\bigskip
			\bigskip
			\bigskip
			\begin{center}
				{\LARGE\bf Diffusion-corrected Autoregressive Fourier Neural Operator for Droplet Evolution Prediction}
			\end{center}
			\medskip
		} \fi
		\bigskip
		
	\begin{abstract}
Predicting droplet evolution in material jetting, or Inkjet Printing (IJP), is essential for maintaining printing quality. However, long-horizon forecasts remain challenging due to error accumulation and the complex coupling of process variables. In this work, we introduce the \textbf{D}iffusion-corrected \textbf{A}uto-\textbf{R}egressive \textbf{F}ourier \textbf{N}eural \textbf{O}perator \textbf{(DiffARFNO)}, a two-stage framework that combines an autoregressive Fourier-MIONet with a conditional Denoising Diffusion Implicit Model (DDIM) corrector. Fourier-MIONet is trained as a coarse predictor and deployed autoregressively for long-horizon forecasting. In the second stage, a DDIM-based conditional corrector refines the coarse prediction within each sliding window through efficient iterative denoising. By combining coarse predictions from Fourier-MIONet with a DDIM corrector that restores fine details, DiffARFNO aims to provide high-fidelity predictions for long-horizon forecasts. Extensive experiments on droplet datasets from ANSYS Fluent demonstrate that DiffARFNO significantly outperforms existing state-of-the-art models. 
	\end{abstract}
			
	\noindent%
	 \vspace{-0.2in}

	\spacingset{1.5} 

\section{Introduction} \label{s:intro}


Inkjet printing (IJP) occupies a distinctive niche within additive manufacturing (AM) as a non-contact, digital, and voxel-wise deposition process that enables fine spatial resolution, multi-material patterning, and scalable area coverage. Unlike extrusion or vat polymerization, inkjet printing ejects picoliter-scale droplets under programmable actuation, enabling rapid prototyping of functional structures (e.g., conductors, dielectrics, biomaterials) and complex ceramic components with minimal tooling and waste \citep{Calvert2001, derby2010inkjet, NGO2018}. These advantages have made inkjet a key AM route for printed electronics, optics, biofabrication, and ceramics. However, the same droplet-level deposition mechanism also introduces substantial modeling and control challenges. IJP is governed by highly transient, nonlinear, and multiscale free-surface dynamics \citep{eggers1997nonlinear, dong2006visualization}. Key fluid properties, including viscosity, density, and surface tension, strongly affect jetting stability, ligament thinning, pinch-off, satellite droplet formation, and wet-on-wet interactions \citep{jang2009influence}. Small variations in these coupled dynamics can lead to unstable ejection, undesired satellite droplets, or inaccurate material placement, thereby directly affecting printing resolution and reliability \citep{sen2021retraction}.

Accurately characterizing these complex droplet dynamics remains challenging because the relevant morphological events occur over micron-scale spatial features and microsecond-scale temporal windows \citep{lohse2022fundamental}. Researchers typically rely on two primary data sources: experimental observation and numerical simulation. High-speed photography combined with precision microscopic optics provides experimental ground truth observations for capturing microsecond-scale droplet phenomena. However, generating such data remains costly. Resolving micron-scale droplet dynamics at microsecond temporal resolution requires ultra-high-speed cameras and precision microscopic optics, which impose substantial equipment and experimental setup costs \citep{derby2010inkjet, castrejon2013future}. Consequently, Computational Fluid Dynamics (CFD) has become a dominant alternative, offering access to internal flow physics that are inaccessible to optical lenses \citep{hirt1981volume, basaran2013nonstandard}. Yet, these classical solvers face their own bottleneck of computational cost. High-fidelity simulations using Volume of Fluid (VOF) methods to resolve thin ligaments and satellites are extremely time-consuming \citep{JIANG2021}. 


To address the computational bottleneck of CFD, researchers have increasingly turned to machine learning models as faster alternatives for droplet simulation and inkjet process analysis. Existing machine learning models for classification and regression have been developed to support task-specific inkjet process analysis. \citet{ogunsanya2021situ} developed a BackPropagation Neural Network (BPNN) droplet-mode classifier for in-situ droplet identification. \citet{li2023multiclass} proposed a multiclass reinforced active learning (MCRAL) framework for identifying pinch-off behavior of droplet evolution. Convolutional Neural Networks-based (CNN-based) monitoring methods have also been introduced for process monitoring~\citep{Zhang31122024}. More recently, time-series deep neural networks have been developed for multi-step process-state prediction and real-time decision-making in additive manufacturing~\citep{tang2025multi}. However, they are not designed to predict the evolution of droplets. Other studies have explored data-driven droplet evolution prediction, including predictive modeling for bioprinting, unsupervised droplet-evolution learning, and tensor time-series analysis for material jetting~\citep{Wu2018, huang2020unsupervised, SEGURA2023103461}. These studies struggled to achieve the high-fidelity long-horizon prediction of droplet evolution. Specifically, this paper aims to solve the following problem: given an initial short sequence of droplet evolution (for example, 50 frames) and material properties (for example, density, viscosity, and surface tension), our goal is to predict the entire future droplet evolution (for example, 550 frames) for different regimes, including ligament thinning, pinch-off dynamics, and satellite droplet formation.

Neural operators provide a suitable foundation for addressing this problem because they learn solution mappings between function spaces and support efficient autoregressive rollouts \citep{li2020fourier,kovachki2023neural}. Nevertheless, one limitation of autoregressive neural operators is that their forecasts gradually blur high-frequency structures under long horizons \citep{mccabe2023towards}. There is a need for a correction mechanism that can recover high-frequency structures while preserving the efficiency of the coarse operator rollouts. Diffusion models can serve as correctors to effectively recover high-frequency details lost by neural operators during autoregressive rollouts~\citep{saharia2022image, Ho2022VDM, wang2023reconstructandgeneratediffusionmodeldetailpreserving}. However, applying diffusion models to long-horizon droplet evolution prediction can be computationally expensive because repeated reverse denoising steps are required~\citep{Ho2020DDPM, song2020denoising}. This suggests a strategy that decouples coarse prediction from fine-detail correction. A natural synthesis is to pair a fast neural operator that produces a stable coarse rollout with a lightweight diffusion corrector that refines high-frequency structures and corrects accumulated errors. Emerging evidence across domains suggests that such hybrids can mitigate smoothing features and improve long-horizon stability with modest computational costs, while simultaneously diminishing error accumulation in autoregressive rollouts and enabling the recovery of high-frequency fine-scale structures \citep{oommen2025integratingneuraloperatorsdiffusion,DONG2025114005}.

We propose \textbf{D}iffusion-corrected \textbf{A}uto-\textbf{R}egressive \textbf{F}ourier \textbf{N}eural \textbf{O}perator (DiffARFNO), a two-stage framework that combines an autoregressive Fourier-MIONet \citep{jiang2024fouriermionet} with a conditional Denoising Diffusion Implicit Model \citep[DDIM,][]{song2020denoising} corrector. In the first stage, the Fourier-MIONet is trained and deployed as the coarse prediction model for autoregressive rollout. In the second stage, a DDIM-based conditional corrector corrects the coarse prediction within each sliding window through efficient iterative denoising. By pairing Fourier-MIONet coarse prediction with a DDIM-based corrector to provide fine details, DiffARFNO aims to provide high-fidelity prediction. Our contributions are summarized as follows:

\begin{itemize}
    \item We formulate inkjet droplet evolution as a long-horizon prediction problem: given the initial sequence of droplet evolution (for example, 50 frames in our setting) and material properties, the task is to predict the entire future sequence of droplet evolution under different dynamics.
    
    
    \item We propose a two-stage framework called DiffARFNO. It combines an autoregressive Fourier-MIONet for efficient coarse droplet evolution prediction with a DDIM corrector to provide fine details. We also introduce a binary mask-based training loss, which further improves prediction accuracy across multiple evaluation metrics.
    
    \item We evaluate the proposed method across three representative droplet-evolution regimes: ligament thinning, pinch-off dynamics, and satellite droplet formation. Compared with FNO-based benchmarks, DiffARFNO achieves lower Mean Absolute Error (MAE) and Mean Squared Error (MSE), higher $R^2$, and superior Intersection over Union (IoU), Peak Signal-to-Noise Ratio (PSNR), and Structural Similarity Index Measure (SSIM), demonstrating improved accuracy in both pixel-wise droplet reconstruction and velocity prediction. 

\end{itemize}


This article is organized as follows. Section~\ref{s:sec2} provides the theoretical background of the main components underlying our proposed model. Section~\ref{sec:methods} presents the proposed methodology in detail. Section~\ref{sec:dataset} describes the droplet dataset from the CFD simulation. Section~\ref{sec:Result} reports the quantitative and qualitative results of the proposed DiffARFNO. Finally, Section~\ref{s:conclusion} concludes the article and outlines future research directions.

\section{Literature Review} \label{s:sec2}
\subsection{Inkjet Droplet Dynamics and Modeling Challenges}

Material jetting (MJ), particularly IJP, is a non-contact and digitally controlled voxel-wise deposition technology in additive manufacturing \citep{Calvert2001, NGO2018,li2021data,yuan2022trends}. By selectively ejecting picoliter-scale liquid-phase materials onto a substrate, IJP enables fine spatial resolution, multi-material patterning, and scalable area coverage, making it an important route for fabricating printed electronics, optics, bio-chips, and complex ceramic components \citep{derby2010inkjet}. 

Despite these advantages, the precision and functional integrity of inkjet-printed parts are highly sensitive to the morphological evolution of ejected droplets \citep{Segura2021}. Droplet formation, trajectory, stability, and breakup are governed by a complex multiphysics interplay among actuation waveforms, nozzle back-pressure, and material properties such as density, viscosity, and surface tension \citep{derby2010inkjet, liu2019experimental, Xu2017, Xu2019}. Dimensionless quantities such as the Ohnesorge number ($\mathit{Oh}$), Weber number ($\mathit{We}$), and Reynolds number ($\mathit{Re}$) are commonly used to characterize jettability windows \citep{liu2019experimental}. However, these static metrics cannot fully describe transient interfacial events during droplet ejection and free-flight evolution~\citep{Xu2017, Xu2019}. When these events are not properly controlled, printing defects such as off-trajectory ejection, splashing, undesired satellite droplets, and inaccurate material placement can occur \citep{Segura2021}. Therefore, predictive models that can anticipate droplet morphology and motion are essential for process monitoring and control in inkjet printing \citep{Wu2018}.

Experimental imaging and numerical simulation have been widely used to study droplet dynamics. High-speed imaging provides direct observation of droplet breakup and satellite formation, but resolving micron-scale features over microsecond time intervals requires costly optical equipment and careful experimental setup \citep{derby2010inkjet, castrejon2013future}. On the other hand, CFD provides a high-fidelity alternative by solving the Navier--Stokes equations and tracking the liquid--gas interface, commonly through the VOF method \citep{hirt1981volume, JIANG2021}. CFD can capture internal transient flow physics and complex topological transitions that are difficult to observe experimentally. However, resolving ligament thinning, pinch-off dynamics, and satellite droplet formation requires dense meshes and very small time steps, making CFD computationally expensive for long-horizon prediction and broad parameter-space exploration. This computational bottleneck motivates the development of fast surrogate models for droplet evolution prediction.

\subsection{Data-Driven Approaches for Inkjet Process}


Data-driven approaches have been explored as efficient alternatives for modeling manufacturing processes \citep{wang2021stressnet, wang2022np, wang2023mvgcn, zhao2025ads, liu2025uni}. Early data-driven studies mainly relied on statistical design and empirical optimization methods to tune inkjet printing parameters~\citep{Couto2018RSM}. Although these methods are useful for identifying favorable operating conditions, they typically focus on scalar process responses and cannot model the entire droplet evolution. Other machine learning studies have addressed task-specific problems in inkjet printing and additive manufacturing, such as droplet classification \citep{ogunsanya2021situ}, pinch-off identification \citep{li2023multiclass}, process optimization \citep{Zhang31122024},  and droplet-parameter prediction~\citep{tang2025multi}. However, these methods mainly characterize the printing process using droplet classes, anomaly indicators, or scalar parameters, rather than predicting the entire droplet evolution. Several studies have attempted to predict droplet evolution. Deep recurrent learning on video datasets has been used to model temporal changes in droplet morphology during jetting~\citep{huang2020unsupervised}. Tensor time series analysis has also been introduced to predict droplet evolution in material jetting \citep{SEGURA2023103461}. Although these approaches move beyond static classification or regression, long-horizon autoregressive rollouts remain challenging due to temporal inconsistency, error accumulation, and fidelity degradation~\citep{mccabe2023towards}. These challenges become more severe during pinch-off dynamics and satellite droplet formation, highlighting the need for a prediction framework that preserves both temporal consistency and fine-scale droplet evolution structures.

\subsection{Neural Operators and Diffusion-Based Refinement}

Unlike conventional regression models that predict a small set of scalar quantities, neural operators learn solution mappings between function spaces, making them suitable for efficiently predicting droplet evolution under different physical conditions governed by parametric PDEs~\citep{li2020fourier, kovachki2023neural, kovachki2024operator}. Fourier Neural Operator (FNO) learns solution operators in the Fourier domain and has become a widely used neural operator architecture~\citep{li2020fourier}. Other operator-learning frameworks, such as DeepONet, approximate nonlinear solution operators by combining encoded input functions with spatial query locations through branch and trunk networks \citep{Lu_2021}. These models provide efficient data-driven approximations for PDE-governed systems. However, autoregressive neural operator forecasts can gradually accumulate errors and blur fine-scale structures, especially under complex dynamics and long prediction horizons \citep{lippe2023pde}. This limitation motivates the use of a corrector module that can correct accumulated prediction errors and restore missing fine-scale details.

Diffusion models provide a complementary mechanism for recovering fine-scale details from coarse neural operator predictions. Denoising diffusion probabilistic models (DDPMs) and score-based generative models learn data distributions through gradual noising and denoising processes \citep{Ho2020DDPM, song2020Score}. Sampling accelerations such as DDIM reduce the cost of reverse diffusion while preserving generation quality, making diffusion-based refinement more computationally tractable \citep{song2020denoising}. In this setting, conditioning the diffusion model on coarse neural operator predictions provides a natural way to guide the denoising process toward high-fidelity corrections. It allows the model to recover missing high-frequency details without generating the entire droplet evolution from scratch. This idea has been demonstrated in image super-resolution and video generation tasks \citep{saharia2022image, Ho2022VDM}, and has recently been extended to computational physics for reconstructing flow fields, solving PDE-related inverse problems, and improving long-horizon physical predictions \citep{shu2023physics, huang2024diffusionpde, KOHL2026108641}.

Taken together, there are no prior studies targeting long-horizon prediction of droplet evolution. The neural operator models can efficiently predict coarse evolution but may lose details under long autoregressive rollouts. Diffusion models provide a complementary correction mechanism for recovering high-frequency structures, but using them as long-horizon generators can be computationally expensive because repeated reverse denoising steps are required~\citep{Ho2020DDPM, song2020denoising, Shih2023ParaDiGMS}. These motivate a two-stage framework for droplet evolution prediction, in which a neural operator generates coarse rollout predictions and a diffusion corrector corrects coarse prediction errors and restores high-frequency structures~\citep{oommen2025integratingneuraloperatorsdiffusion, DONG2025114005}.

\section{Methods} \label{sec:methods}
\begin{figure}[!htb]
    \centering
    \includegraphics[width=1\linewidth]{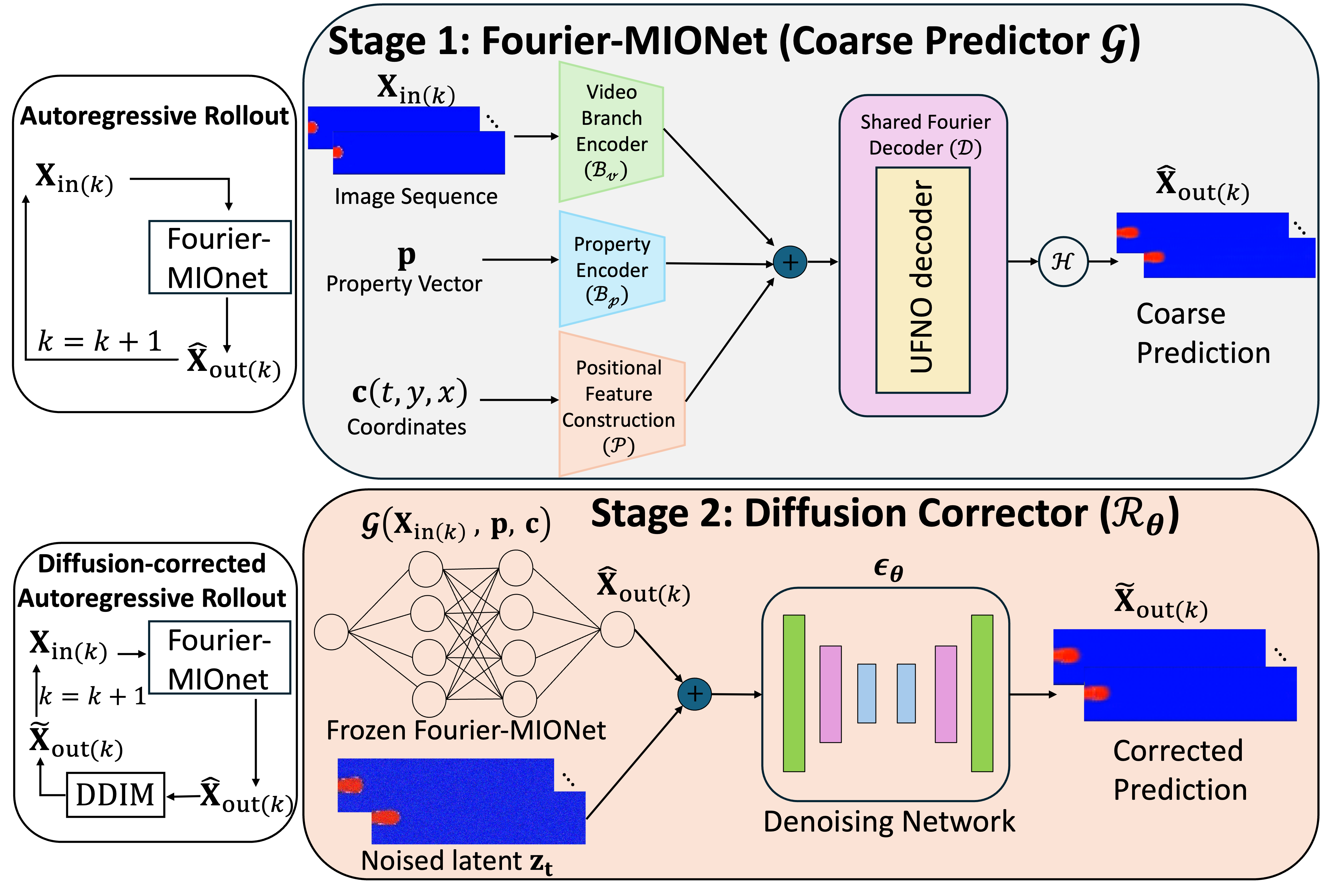} \vspace{-0.3in}
    \caption{Overview of the proposed DiffARFNO framework. In \textbf{Stage~1}, a Fourier-MIONet backbone serves as a coarse autoregressive predictor. At rollout step $k$, the input history $\mathbf{X}_{\mathrm{in}(k)}$, the property vector $\mathbf{p}$, and the coordinate grid $\mathbf{c}(t,y,x)$ are encoded and fused within a shared UFNO decoder, producing a coarse prediction $\widehat{\mathbf{X}}_{\mathrm{out}(k)}$. In \textbf{Stage~2}, a conditional diffusion model refines the coarse prediction, conditioned on the frozen backbone output, to correct accumulated errors and recover fine-scale structures. The refined segment $\widetilde{\mathbf{X}}_{\mathrm{out}(k)}$ is then fed back as updated history for subsequent autoregressive rollout, and the sequence of refined frames across rollout steps is assembled to form the final long-horizon prediction.
}\vspace{-0.15in}
    \label{fig:framework}
\end{figure}

Given an initial image sequence $\mathbf{X}_{1:L}=\{\mathbf{X}_1,\mathbf{X}_2,\cdots, \mathbf{X}_L\}$, where $\mathbf{X}_i\in\mathbb{R}^{3\times H\times W}$ denotes the $i$th RGB image frame of size $H\times W$, our goal is to predict the future image sequence $\mathbf{X}_{{L+1}:S}$. In our case study in Section~\ref{sec:Result}, we have $L=50$ and $S=600$. Instead of building a model to predict from $\mathbf{X}_{1:L}$ to the future $\mathbf{X}_{{L+1}:S}$ in one step, we aim to build an autoregressive forecasting framework for long-horizon droplet evolution as shown in Figure~\ref{fig:framework}. The framework contains two stages: at Stage 1 in Section~\ref{sec:backbone}, a neural operator backbone is applied autoregressively to produce coarse long-horizon predictions; at Stage 2 in Section~\ref{sec:diffusion_refine}, a DDIM corrector then corrects residual errors and restores fine-scale structures on top of the outputs from the frozen backbone.

\subsection{\emph{Stage 1: Fourier-MIONet Backbone}}
\label{sec:backbone}

Our coarse predictor follows the structure of Fourier-MIONet~\citep{jiang2024fouriermionet} to handle multi-source inputs with a branch--trunk operator structure as shown in the top panel of Figure~\ref{fig:framework}.  The Fourier-MIONet has three inputs: (1) the image sequence input $\mathbf{X}_{\mathrm{in}(k)}$; (2) the static property vector $\mathbf{p}$ (for example, density, viscosity, and surface tension in our case); (3) a normalized coordinate grid $\mathbf{c}(t,y,x)\in[-1,1]^3$. $\mathbf{X}_{\mathrm{in}(k)}$ is processed by the video branch encoder  $\mathcal{B}_v$, $\mathbf{p}$ is processed by the property encoder $\mathcal{B}_p$, and $\mathbf{c}(t,y,x)$ is processed by the positional feature construction/projection $\mathcal{P}$. Further, these processed representations are fused and decoded by a shared Fourier-based decoder $\mathcal{D}$, which is a UFNO decoder. A point-wise head $\mathcal{H}$, a multilayer perceptron (MLP), maps decoded features back to the output $\widehat{\mathbf{X}}_{\mathrm{out}(k)}$. The coarse prediction can be represented as 
\begin{equation}
\widehat{\mathbf{X}}_{\mathrm{out}(k)}
=
\mathcal{H}\!\left(
\mathcal{D}\!\left(
\mathrm{Concat}\!\left[
\mathcal{B}_v(\mathbf{X}_{\mathrm{in}(k)}),\,
\mathrm{Broadcast}\!\big(\mathcal{B}_p(\mathbf{p})\big),\,
\mathcal{P}(\mathbf{c})
\right]
\right)
\right)
:=
\mathcal{G}\!\left(\mathbf{X}^{(k)}_{\mathrm{in}},\,\mathbf{p},\,\mathbf{c}\right),
\label{eq:coarse_operator}
\end{equation}
where $\mathrm{Concat}[\cdot]$ concatenates feature tensors along the channel dimension, $\mathrm{Broadcast}(\cdot)$ replicates the condition code across the spatiotemporal grid, and $\mathcal{G}$ denotes the Fourier-MIONet-based coarse predictor.

After we trained Fourier-MIONet,  our autoregressive rollout works in the following way:  $\mathbf{X}_{\text{in}(0)} \to \texttt{Fourier-MIONet} \to \widehat{\mathbf{X}}_{\text{out}(0)}:=\mathbf{X}_{\text{in}(1)}\to \texttt{Fourier-MIONet} \to \widehat{\mathbf{X}}_{\text{out}(1)}:=\mathbf{X}_{\text{in}(2)}\to \cdots\to \widehat{\mathbf{X}}_{\text{out}(K)}$. Specifically, we have

\noindent\textbf{For $k=0$:}
\begin{equation}
\mathbf{X}_{\text{in}(k)} = \mathbf{X}_{1:L},
\qquad
\widehat{\mathbf{X}}_{\text{out}(k)} = \widehat{\mathbf{X}}_{L+1:2L}.
\label{eq:io_k0}
\end{equation}
\noindent\textbf{For $k>0$:}
\begin{equation}
\mathbf{X}_{\text{in}(k)} = \widehat{\mathbf{X}}_{[(L-\Delta)k+1]:\,[(L-\Delta)k+L]},
\quad
\widehat{\mathbf{X}}_{\text{out}(k)} = \widehat{\mathbf{X}}_{[(L-\Delta)k+L+1]:\,[(L-\Delta)k+2L]}.
\label{eq:io_kpos}
\end{equation}
Note that $\widehat{\mathbf{X}}$ represents prediction, and the rollout step is defined as $k\in\{0,1,2,\cdots,K\}$. Since the model performs an autoregressive rollout in fixed-length segments during prediction, mismatches can occur between consecutive segments. If these windows are concatenated directly, the mismatch can cause droplet shaking at the window boundaries. To mitigate this, a linear blending is used to softly align the first $\Delta$ frames of the current prediction with the last $\Delta$ frames of the previous prediction, improving continuity and rollout stability since the fused frames are fed back as context to form $\mathbf{X}_{\text{in}(k+1)}$. Specifically, we linearly fuse the overlapping region by updating the first $\Delta$ frames of the current window using the last $\Delta$ frames of the previous window. For $j\in\{1,\dots,\Delta\}$ and  $k>0$
\begin{equation}
\widehat{\mathbf{X}}_{\text{out}(k)}[j]
\leftarrow
w_j\,\widehat{\mathbf{X}}_{\text{out}(k-1)}[L-\Delta+j]
+
(1-w_j)\,\widehat{\mathbf{X}}_{\text{out}(k)}[j],
\label{eq:blend_update}
\end{equation}
where $\widehat{\mathbf{X}}_{\text{out}(k)}[j]\in\mathbb{R}^{3\times H\times W}$ denotes the $j$-th frame in the step-$k$ output window, and
\begin{equation}
w_j=\frac{\Delta-j}{\Delta-1}.
\label{eq:blend_weight}
\end{equation}
After fusion, the stitched sequence is used to form the next input window $\mathbf{X}_{\text{in}(k+1)}$ under Eqs.~\eqref{eq:io_k0}--\eqref{eq:io_kpos}.




To complement pixel-wise reconstruction and promote structural fidelity of the predicted droplet evolution, we incorporate a mask-aligned supervision term during training. Given an RGB image $\mathbf{X}\in\mathbb{R}^{3\times H\times W}$, we first convert it to a grayscale intensity map
\begin{equation}
G(y,x)
=
\sum_{c=1}^{3}\omega_c\,\mathbf{X}_c(y,x),
\qquad (\omega_1,\omega_2,\omega_3)=(0.299,\,0.587,\,0.114),
\label{eq:gray}
\end{equation}
and then obtain the binary mask as
\begin{equation}
\mathcal{M}(\mathbf{X})(y,x)
=
\mathbbm{1}[G(y,x)\ge \tau^{\star}(G)],
\label{eq:mask_bin_singleop}
\end{equation}
where $\tau^{\star}(G)$ is the Otsu threshold \citep{Otsu1979} computed from $G$. $\mathbbm{1}[\cdot]$ denotes the indicator function, which equals 1 when the condition inside the brackets is true and 0 otherwise. The same $\mathcal{M}(\cdot)$ is applied to both predictions and ground truth in the mask loss. For each training step $k$, we minimize
\begin{equation}
\mathcal{L}_{\mathrm{Stage 1}}
=
\lambda_{\mathrm{rgb}}
\big\|
\widehat{\mathbf{X}}_{\text{out}(k)}-\mathbf{X}_{\text{out}(k)}
\big\|_2^2
+
\lambda_{\mathrm{mask}}
\big\|
\mathcal{M}(\widehat{\mathbf{X}}_{\text{out}(k)})-\mathcal{M}(\mathbf{X}_{\text{out}(k)})
\big\|_1,
\label{eq:stage1_loss}
\end{equation}
where $\lambda_{\mathrm{rgb}}=1$ and $\lambda_{\mathrm{mask}}=0.5$ are the weights of the RGB and mask losses, respectively. The weights were selected empirically using validation-set performance and kept fixed across all experiments.

\subsection{\emph{Stage 2: Conditional Diffusion Correction}}
\label{sec:diffusion_refine}

In the second stage, the coarse predictor $\mathcal{G}$ is fixed and only the diffusion corrector is trained. Specifically, we sample Gaussian noise $\boldsymbol{\epsilon}\sim\mathcal{N}(\mathbf{0},\mathbf{I})$ and form
\begin{equation}
\mathbf{z}_t
=
\sqrt{\bar{\alpha}_t}\,
\mathbf{X}_{\text{out}(k)}
+
\sqrt{1-\bar{\alpha}_t}\,
\boldsymbol{\epsilon},
\label{eq:stage2_forward_noise}
\end{equation}
where $\{\bar{\alpha}_t\}$ is computed from a linear beta schedule with $\beta_t$ linearly increasing from $10^{-4}$ to $2\times10^{-2}$ over 200 diffusion steps \citep{Ho2020DDPM}. For each rollout step $k$, the corrector is implemented by a conditional denoiser
$\boldsymbol{\epsilon}_\theta(\cdot)$ and a DDIM sampling trajectory conditioned on
the coarse prediction $\widehat{\mathbf{X}}_{\text{out}(k)}$. We first form the predicted clean estimate
\begin{equation}
\widetilde{\mathbf{X}}^{clean}_{\text{out}(k)}
=
\frac{\mathbf{z}_t-\sqrt{1-\bar{\alpha}_t}\,
\boldsymbol{\epsilon}_\theta\!\left(\mathbf{z}_t, t \mid \widehat{\mathbf{X}}_{\text{out}(k)}\right)}
{\sqrt{\bar{\alpha}_t}}.
\label{eq:ddim_x0_pred_tilde}
\end{equation}
Then one DDIM reverse step generates $\mathbf{z}_{t-1}$ from $\mathbf{z}_t$ as
\begin{equation}
\mathbf{z}_{t-1}
=
\sqrt{\bar{\alpha}_{t-1}}\,
\widetilde{\mathbf{X}}^{clean}_{\text{out}(k)}
+
\sqrt{1-\bar{\alpha}_{t-1}-\sigma_t^2}\;
\boldsymbol{\epsilon}_\theta\!\left(\mathbf{z}_t, t \mid \widehat{\mathbf{X}}_{\text{out}(k)}\right)
+
\sigma_t\,\boldsymbol{\eta},
\qquad
\boldsymbol{\eta}\sim\mathcal{N}(\mathbf{0},\mathbf{I}).
\label{eq:ddim_update_tilde}
\end{equation}
Here, $\sigma_t$ controls the amount of stochastic noise injected at the DDIM reverse step, and $\boldsymbol{\eta}$ is an independent standard Gaussian noise term sampled from $\mathcal{N}(\mathbf{0},\mathbf{I})$. In our implementation, we use the deterministic DDIM update with $\sigma_t=0$ during sampling so the last stochastic noise term is removed during inference. At inference, we initialize $\mathbf{z}_t$ at the final diffusion step as Gaussian noise and iterate
Eq.~\eqref{eq:ddim_update_tilde} backward to $t=0$, yielding a deterministic mapping from
$\widehat{\mathbf{X}}_{\text{out}(k)}$ to the refined output.
Accordingly, we denote 
\begin{equation}
\widetilde{\mathbf{X}}_{\text{out}(k)}
=
\mathcal{R}_\theta\!\left(\widehat{\mathbf{X}}_{\text{out}(k)}\right),
\qquad k=0,1,2,\dots
\label{eq:R_step_final}
\end{equation}
where $\mathcal{R}_\theta$ denotes the full diffusion corrector, including the noisy latent initialization, the conditioning on the coarse prediction, and the DDIM reverse sampling trajectory.

For each rollout step $k$, our two-stage forecaster first processes inputs with the coarse predictor $\mathcal{G}$ from Section~\ref{sec:backbone} and then with the refinement operator $\mathcal{R}_\theta$ as
\begin{equation}
\widetilde{\mathbf{X}}_{\text{out}(k)}
=
\mathcal{R}_\theta\!\Big(
\mathcal{G}\!\left(\mathbf{X}_{\text{in}(k)},\,\mathbf{p},\,\mathbf{c}\right)
\Big),
\qquad k=0,1,2,\dots
\label{eq:overall_twostage_inline}
\end{equation}
where $\widetilde{\mathbf{X}}_{\text{out}(k)}$ denotes the refined output sequence at rollout step $k$. It is obtained by applying the refinement operator $\mathcal{R}_\theta$ to the coarse prediction generated by $\mathcal{G}$. This refined output is then used as the input for the next rollout step.

For each training step $k$, we minimize the MSE between the Gaussian noise added to the target sequence and the noise predicted by the denoising network.
\begin{equation}
\mathcal{L}_{\mathrm{stage2}}
=
\mathbb{E}_{\mathbf{X}_{\mathrm{out}(k)},\,t,\,\boldsymbol{\epsilon}}
\left[
\left\|
\boldsymbol{\epsilon}
-
\boldsymbol{\epsilon}_{\theta}
\left(
\mathbf{z}_t, t
\,\middle|\,
\widehat{\mathbf{X}}_{\mathrm{out}(k)}
\right)
\right\|_2^2
\right],
\label{eq:stage2_loss}
\end{equation}
where $\mathbb{E}[\cdot]$ denotes the expectation over training target sequences, randomly sampled diffusion timesteps, and Gaussian noise samples. This loss encourages $\boldsymbol{\epsilon}_{\theta}$ to estimate the noise so that the clean sequence can be recovered during the DDIM reverse process.


\section{Dataset Description}\label{sec:dataset}

\begin{figure}[!htb]\vspace{-0.1in}
\centering
\includegraphics[width=\linewidth]{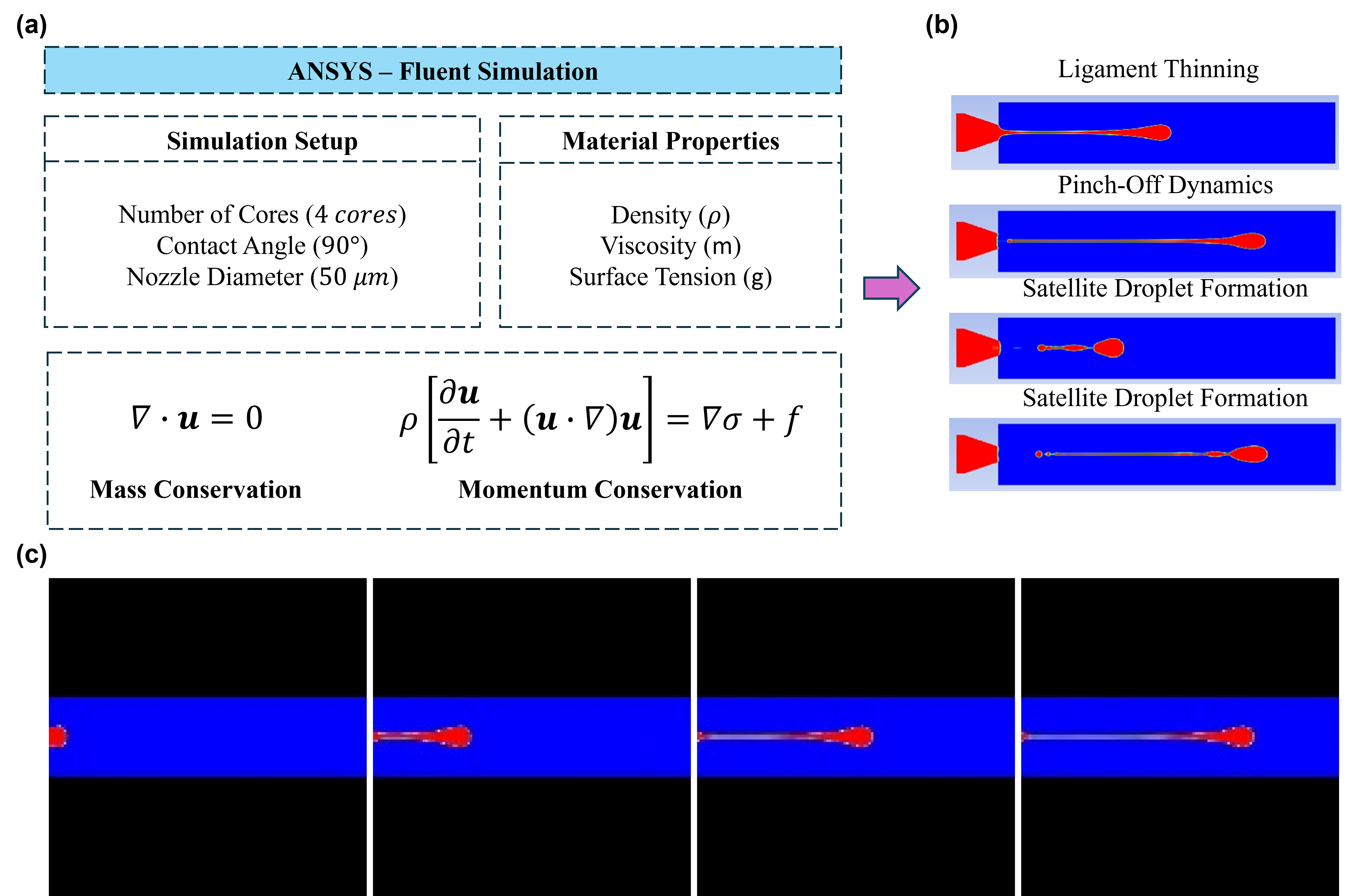}
\caption{Simulated droplet evolution dataset and final training samples.
(a) ANSYS Fluent simulation setup, including computational parameters, fluid material properties, and the governing incompressible Navier–Stokes equations for mass and momentum conservation.
(b) Example snapshots of the droplet formation process before preprocessing, showing representative regimes including ligament thinning, pinch-off, and satellite droplet formation. (c) Example snapshots of the processed ligament thinning regime.}\vspace{-0.15in}
\label{fig:mat_props}
\end{figure}


\subsection{CFD Simulation}
All videos used in this work are generated via a high-fidelity CFD simulation pipeline. The droplet evolution is simulated in ANSYS Fluent \citep{ansys_fluent_user}. Figure~\ref{fig:mat_props}(a) shows the CFD simulation settings and the material properties used to define each simulation case. The Navier-Stokes equations govern the physical system, enforcing mass and momentum conservation across the liquid-gas interface. The fluid phases are assumed to be viscous, axisymmetric, and incompressible \citep{huang2020unsupervised}. To explicitly track the complex topological transitions of the droplet interface over time, such as ligament thinning, pinch-off dynamics, and satellite droplet formation, the VOF method is adopted to capture the morphology and position of the liquid phase \citep{hirt1981volume, JIANG2021}.

\begin{table}[!htbp]
\centering
\caption{Material property ranges used for simulation.}
\label{tab:mat_props}
\begin{tabular}{lcc}
\hline
\textbf{Material property} & \textbf{Low level} & \textbf{High level} \\
\hline
Density (kg/m$^3$)              & 800     & 8000  \\
Viscosity (kg/(m$\cdot$s))      & 0.0005  & 0.15  \\
Surface tension (dyn/cm)        & 50      & 80    \\
\hline
\end{tabular}
\end{table}

To accurately solve the pressure-velocity coupling, the simulations employ a pressure-based solver utilizing a fractional step scheme, least-squares cell-based gradient evaluation, a pressure staggering option, and the Quadratic upstream interpolation for convective kinematics (QUICK) scheme for spatial discretization across quadrilateral and hexahedral meshes \citep{huang2020unsupervised}. The physical nozzle diameter is set to 50~$\mu$m with a contact angle of $90^\circ$. The piezoelectric actuation driving the droplet ejection is computationally modeled by defining the inlet velocity boundary condition as a trapezoidal waveform, characterized by specific rise, dwell, and fall times. By varying material property settings in the CFD runs, we obtain diverse simulated droplet videos spanning different fluid regimes. To efficiently sample the parameter space, a Latin hypercube design using the min-max criterion was utilized to generate distinct combinations of ink parameters \citep{huang2020unsupervised}. The ranges of material properties explored in the simulation—specifically density, viscosity, and surface tension—are summarized in Table~\ref{tab:mat_props}. 


\subsection{Data Curation and Preprocessing}

Following the CFD simulation, dataset curation and image preprocessing are performed to prepare the raw simulation outputs for deep learning. The raw simulation frames are output at a resolution of 1056$\times$512 pixels and contain broad background regions that do not contribute to the droplet dynamics. Figure~\ref{fig:mat_props}(b) illustrates simulation results from distinct droplet-ejection regimes, including ligament thinning, pinch-off dynamics, and satellite droplet formation. These diverse regimes provide a comprehensive dataset for training and evaluating the droplet evolution prediction models. To reduce computational overhead and eliminate irrelevant background information, each video frame is first cropped to a region of interest (ROI) that bounds the droplet ejection trajectory. The cropped ROI frames are then resized to 128$\times$32 pixels. 

For temporal curation, we retain only videos that capture a complete droplet evolution sequence and remove incomplete or failed simulation runs. This process results in a final curated set of 212 simulated videos. The cropped ROI frames are padded to obtain uniform square frames of 128$\times$128 pixels. Snapshots of the processed ligament thinning regime data are shown in Figure~\ref{fig:mat_props}(c). This standardizes the spatial dimensions for efficient Fourier-domain operations in the neural operator models considered in this study. All RGB pixel intensities are normalized to 0--1.

Each curated video consists of 600 RGB frames, with each frame of size $128 \times 128$. Therefore, each video is represented as a tensor of size $600 \times 128 \times 128 \times 3$. In addition, each video is paired with a static condition vector $\mathbf{p}$ representing its material properties, including density, viscosity, and surface tension. These auxiliary continuous variables are normalized to a 0--1 range using min--max scaling to improve numerical stability when fusing property embeddings with the spatiotemporal coordinate grid during operator encoding.

\section{Results} \label{sec:Result}
\subsection{Experimental Setting}

We randomly split the 212 simulated videos at the video level into 149 training videos (70\%), 21 validation videos (10\%), and 42 testing videos (20\%). All reported metrics are computed on the held-out test set, with model selection and hyperparameter tuning performed using the validation set. An early-stopping strategy is applied based on the validation PSNR to reduce overfitting. Specifically, after each epoch, the model is evaluated on the validation set, and training is stopped if the validation PSNR improvement remains smaller than $10^{-3}$ for eight consecutive epochs. The key training hyperparameters and implementation settings are in Appendix~\ref{appendix: hyperparameters}.

We compare the proposed method with the following representative benchmark models:

\begin{itemize}
    \item Fourier-MIONet~\citep{jiang2024fouriermionet}: a multiple-input neural operator that incorporates Fourier-domain representations for spatiotemporal prediction.
    \item PIANO~\citep{cao2025physicsinformedattentionenhancedfourierneural}: a physics-informed and attention-enhanced Fourier neural operator designed to improve fluid-dynamics modeling.
    \item UFNO~\citep{wen2022ufnoenhancedfourier}: a U-shaped Fourier neural operator that improves multiscale feature extraction through an encoder--decoder structure.
    \item FNO~\citep{li2020fourier}: a foundational neural operator that learns mappings between function spaces using Fourier-domain spectral convolution.
    \item GeoFNO~\citep{li2023fourier}: a geometry-aware Fourier neural operator that extends operator learning to irregular or geometry-dependent domains.
\end{itemize}

All benchmark models are trained using the same dataset split, preprocessing pipeline, input--output window configuration, and the same autoregressive prediction protocol as DiffARFNO. For robustness, each model is independently trained and evaluated over ten repeated runs with different random seeds, and the reported results are summarized from these repeated experiments. All training and testing are conducted on a server equipped with eight NVIDIA A100 80GB GPUs.

\subsection{\emph{Quantitative Evaluation on Test Set}} \label{s:methods.2}

We evaluate the proposed DiffARFNO and benchmark models on the test set using six metrics: MSE, MAE, PSNR, SSIM, $R^2$, and IoU. MSE and MAE quantify pixel-wise reconstruction errors, PSNR and SSIM measure image-level fidelity and structural similarity, $R^2$ evaluates goodness-of-fit, and IoU assesses binary mask overlap between predicted and ground truth droplet regions. All metrics are computed over the predicted frames in the test videos.

\newcolumntype{L}[1]{>{\raggedright\arraybackslash}p{#1}}

\begin{table}[t]
\centering
\caption{Test set performance reported as mean and standard deviation (std). Arrows indicate the preferred direction; best mean (and best std, when applicable) per column are in \textbf{bold}.}
\label{tab:test_metrics}
\small
\setlength{\tabcolsep}{6pt}
\renewcommand{\arraystretch}{1.12}

\begin{tabular}{L{2.9cm} c
                S[table-format=1.4]
                S[table-format=1.4]
                S[table-format=2.4]
                S[table-format=1.4]
                S[table-format=1.4]
                S[table-format=1.4]}
\toprule
\textbf{Model} &
\textbf{Stat} &
{\textbf{MSE}$\downarrow$} &
{\textbf{MAE}$\downarrow$} &
{\textbf{PSNR}$\uparrow$} &
{\textbf{SSIM}$\uparrow$} &
{\textbf{$R^2$}$\uparrow$} &
{\textbf{IoU}$\uparrow$} \\
\midrule
\addlinespace[1pt]

DiffARFNO
& Mean & \textbf{0.0003} & \textbf{0.0033} & \textbf{34.7104} & \textbf{0.9763} & \textbf{0.9945} & \textbf{0.8711} \\
& Std  & \textbf{0.0000} & \textbf{0.0001} & \textbf{0.0243}  & \textbf{0.0017} & \textbf{0.0001} & \textbf{0.0019} \\
\addlinespace[3pt]\midrule\addlinespace[3pt]

Fourier-MIONet
& Mean & 0.0004 & 0.0046 & 34.4669 & 0.9555 & 0.9939 & 0.8394 \\
& Std  & 0.0001 & \textbf{0.0001} & 0.3820 & 0.0058 & 0.0005 & 0.0155 \\
\addlinespace[3pt]\midrule\addlinespace[3pt]

PIANO
& Mean & 0.0007 & 0.0072 & 32.3504 & 0.9103 & 0.9905 & 0.7965 \\
& Std  & 0.0001 & \textbf{0.0001} & 0.1453 & 0.0041 & 0.0002 & 0.0038 \\
\addlinespace[3pt]\midrule\addlinespace[3pt]

UFNO
& Mean & 0.0014 & 0.0042 & 29.2835& 0.9756 & 0.9804 & 0.7264 \\
& Std  & 0.0003 & 0.0005 & 1.1086 & 0.0056 & 0.0043 & 0.0387 \\
\addlinespace[3pt]\midrule\addlinespace[3pt]

FNO
& Mean & 0.0011 & 0.0070 & 30.4833 & 0.9384 & 0.9846 & 0.7202 \\
& Std  & 0.0001 & \textbf{0.0001} & 0.0686 & 0.0019 & 0.0003 & 0.0058 \\
\addlinespace[3pt]\midrule\addlinespace[3pt]

GeoFNO
& Mean & 0.0006 & 0.0064 & 33.2020 & 0.9305 & 0.9822 & 0.6987 \\
& Std  & 0.0002 & 0.0005 & 0.1183 & 0.0040 & 0.0002 & 0.0183 \\
\bottomrule
\end{tabular}
\end{table}

Table~\ref{tab:test_metrics} reports the test set performance of DiffARFNO and five benchmark models, including Fourier-MIONet, PIANO, UFNO, FNO, and GeoFNO. The reported mean and standard deviation (std) are computed over ten repeated independent runs. DiffARFNO achieves the best mean performance across all evaluation metrics, with MSE of 0.0003, MAE of 0.0033, PSNR of 34.7104, SSIM of 0.9763, $R^2$ of 0.9945, and IoU of 0.8711. Fourier-MIONet is the second-best model, achieving comparable reconstruction accuracy but lower structural and segmentation fidelity, especially in IoU. The remaining benchmarks show larger degradation in either pixel-wise reconstruction or mask overlap, with IoU values ranging from 0.6987 for GeoFNO to 0.7965 for PIANO. These results indicate that DiffARFNO improves both image reconstruction and droplet localization on test sequences. 

In terms of computational time, conventional CFD simulation requires approximately \textbf{4 hours} to generate a complete droplet-evolution sequence; DiffARFNO produces the corresponding prediction in only about \textbf{40 seconds}.

\subsection{\emph{Qualitative Evaluation on RGB Images and Binary Masks}}
\label{sec:qual_results}

To complement the quantitative evaluation in Table~\ref{tab:test_metrics}, we present a qualitative evaluation on three representative regimes:  Section~\ref{subsubsec: ligament}: ligament thinning; Section~\ref{subsubsec: pinch}: pinch-off dynamics; and Section~\ref{subsubsec: satellite}: satellite droplet formation. More results are provided in Appendix~\ref{appendix: binary}. 
\begin{figure}[!htbp]
\centering
\includegraphics[width=\linewidth]{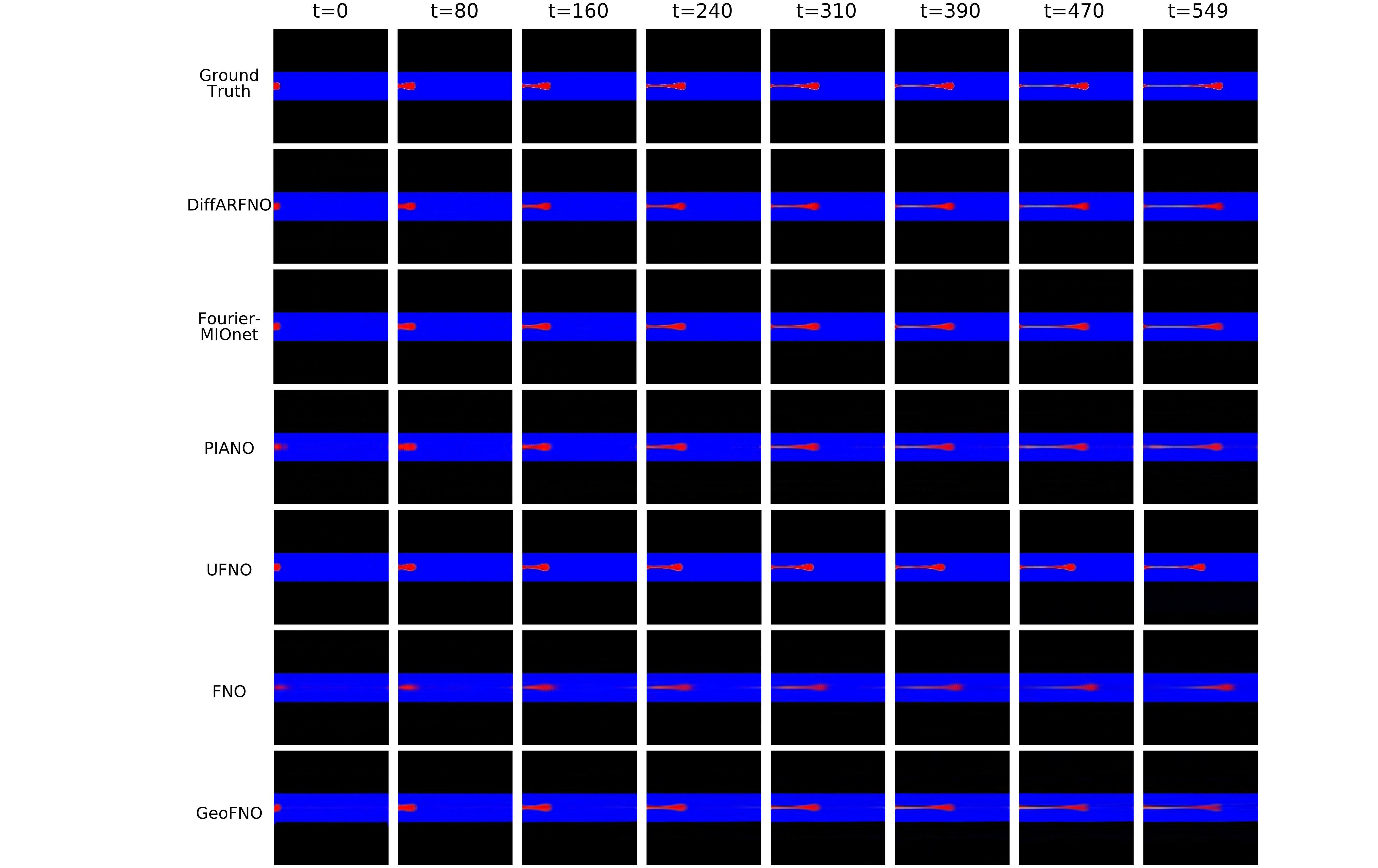}\vspace{-0.05in}
\caption{RGB predictions of \textbf{ligament thinning} regime.} \vspace{-0.25in}
\label{fig:qual_sample41}
\end{figure}

\subsubsection{\emph{Ligament Thinning}} \label{subsubsec: ligament}
\begin{figure}[!htbp]
\centering
\includegraphics[width=\linewidth]{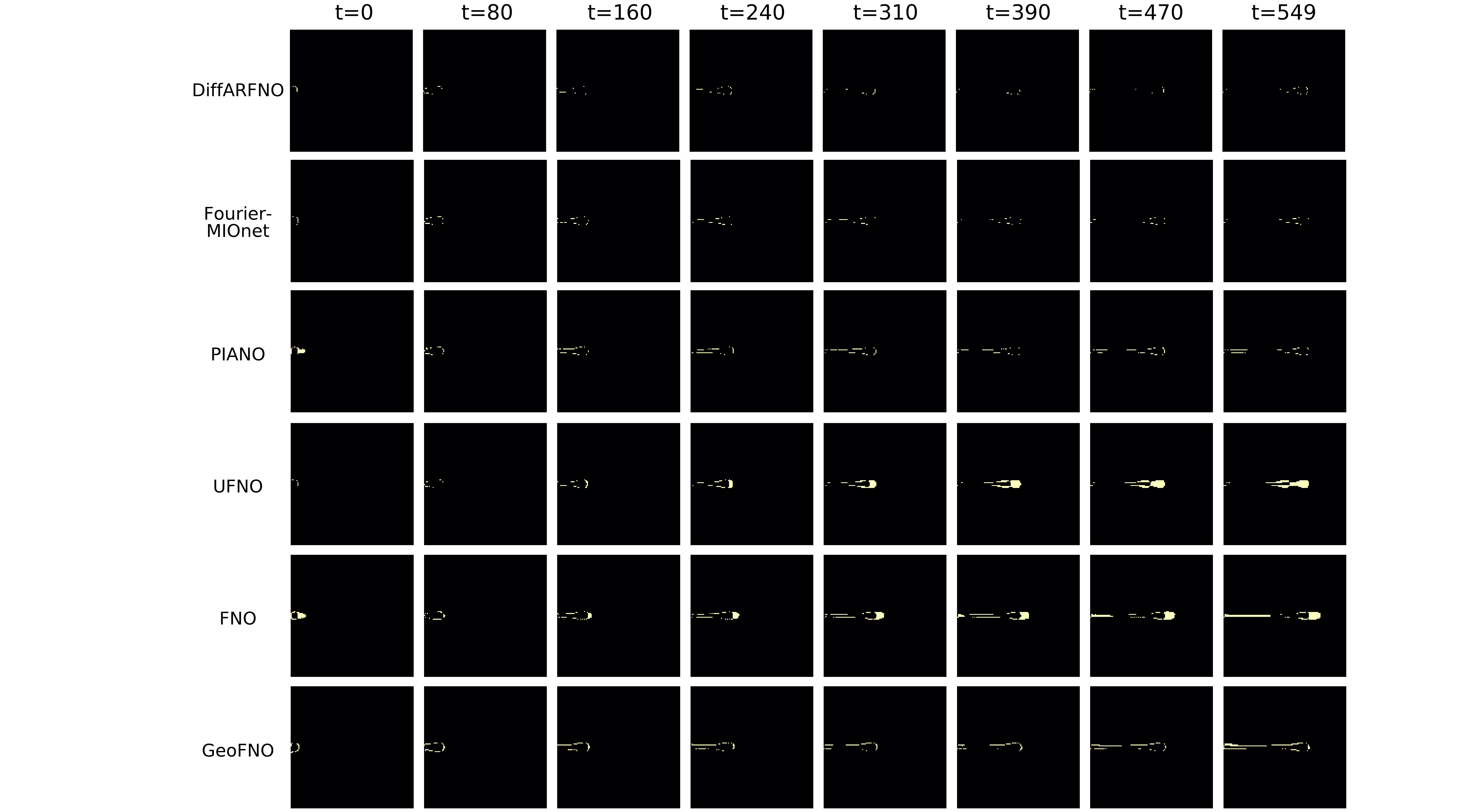}\vspace{-0.05in}
\caption{The error map of binary masks for \textbf{ligament thinning} regime. The error map is calculated between the prediction mask and the ground truth mask.} \vspace{-0.15in}
\label{fig:qual_sample41_error}
\end{figure}

Figure~\ref{fig:qual_sample41} shows the RGB predictions for \textbf{ligament thinning} regime, in which the droplet ligament undergoes gradual elongation and thinning while remaining topologically connected throughout the sequence. This regime shows the model fidelity, as accurate prediction requires preserving extremely slender structures and maintaining stable droplet thickness without introducing breakup or excessive smoothing. Within each RGB predictions panel, rows are ordered from top to bottom as follows: Ground Truth, DiffARFNO, Fourier-MIONet, PIANO, UFNO, FNO, and GeoFNO. The error maps in Figure~\ref{fig:qual_sample41_error} are computed based on the predicted and ground truth binary masks and rows are ordered from top to bottom as follows: DiffARFNO, Fourier-MIONet, PIANO, UFNO, FNO, and GeoFNO.

DiffARFNO exhibits the closest visual agreement with the ground truth in RGB sequences. It maintains a clear neck over time, with minimal boundary shifts and minor thickness changes, indicating strong capability in capturing fine-scale dynamics. Fourier-MIONet reliably reproduces the global elongation trend and overall droplet morphology; however, mild smoothing becomes evident as the neck thins, resulting in less clear boundaries. PIANO produces visually plausible evolution in early and intermediate stages, but errors in neck thickness become more noticeable at later time steps, suggesting reduced robustness during the autoregressive process.

In contrast, FNO tends to over-smooth the thin ligament and introduces local discontinuities in the predicted evolution, leading to a less physically consistent representation of the droplet morphology. This behavior suggests that standard neural operator formulations may struggle to preserve the continuity of the ligament thinning regime. Moreover, UFNO has a spatiotemporal mismatch between the predicted morphology and the ground truth. GeoFNO improves global morphology preservation relative to UFNO and FNO, particularly in maintaining the overall ligament alignment, but still exhibits boundary and thickness errors compared with DiffARFNO.


\begin{figure}[!htbp]
\centering
\includegraphics[width=\linewidth]{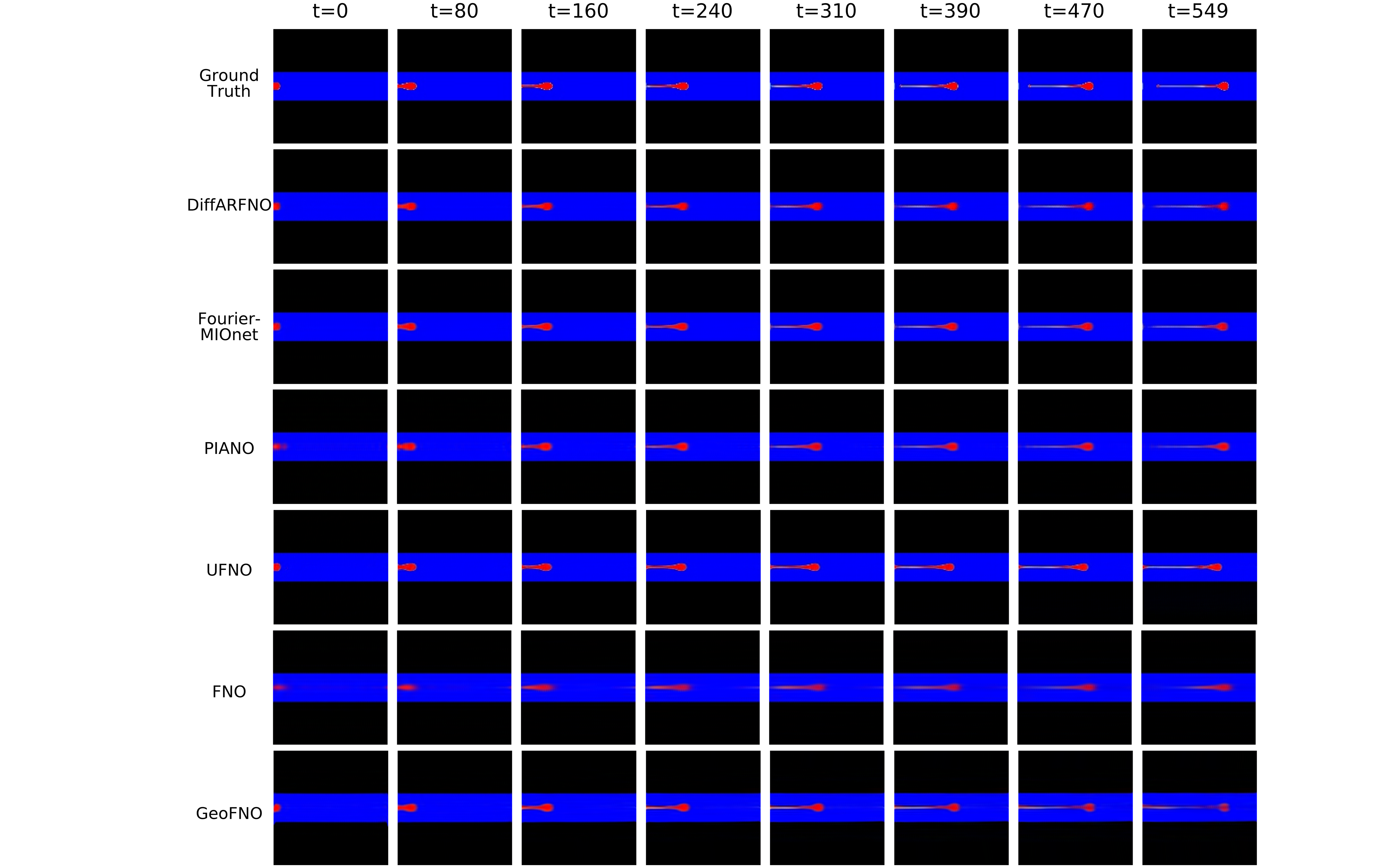}\vspace{-0.05in}
\caption{RGB predictions of \textbf{pinch-off dynamics} regime.} \vspace{-0.15in}
\label{fig:qual_sample39}
\end{figure}

\subsubsection{\emph{Pinch-Off Dynamics}} \label{subsubsec: pinch}

\begin{figure}[!htbp]
\centering
\includegraphics[width=\linewidth]{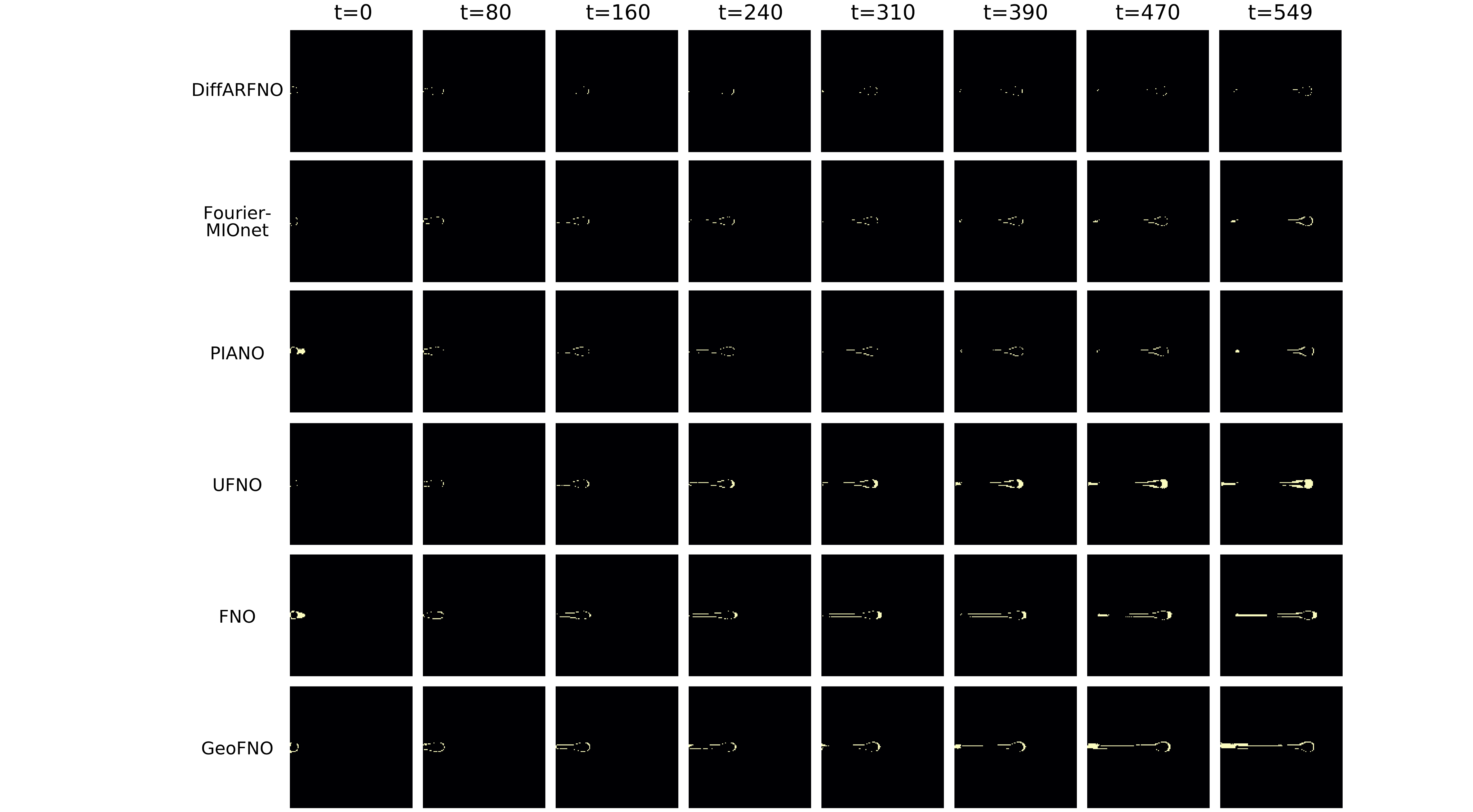}\vspace{-0.05in}
\caption{The error map of binary masks for \textbf{pinch-off dynamics} regime. The error map is calculated between the prediction mask and the ground truth mask.} \vspace{-0.15in}
\label{fig:qual_sample39_error}
\end{figure}

Figure~\ref{fig:qual_sample39} corresponds to a \textbf{pinch-off dynamics} regime, in which the elongated ligament develops a thinning neck that approaches breakup. The error maps in Figure~\ref{fig:qual_sample39_error} are computed based on the predicted and ground truth masks. Accurately resolving this process requires capturing severe neck thinning, the localization of the pinch-off point, and the temporal consistency of the droplet morphology. DiffARFNO shows the closest agreement with the ground truth in RGB sequences. It preserves the ligament thinning and maintains a clear neck region over time, indicating strong capability in modeling pinch-off dynamics.






In comparison, Fourier-MIONet captures the global droplet evolution and overall elongation trend, but mild smoothing is visible near the neck region, leading to less clear mask boundaries around the breakup point. PIANO produces plausible global morphology, but its predictions show larger boundary variations and reduced sharpness when the neck becomes highly slender, suggesting limited robustness to morphology changes.

For UFNO and FNO, the degradation is more noticeable. FNO tends to blur or weaken the thinning ligament, which leads to a less accurate mask. UFNO fails to predict the occurrence of droplet breakup. As a result, UFNO and FNO may exhibit a mismatch relative to the ground truth near the pinch-off region. GeoFNO shows noticeable spatial misalignment and severe smoothing of the ligament. The predicted morphology fails to match the ground truth evolution and neck morphology, indicating limited sensitivity to pinch-off dynamics.


\begin{figure}[!htbp]
\centering
\includegraphics[width=\linewidth]{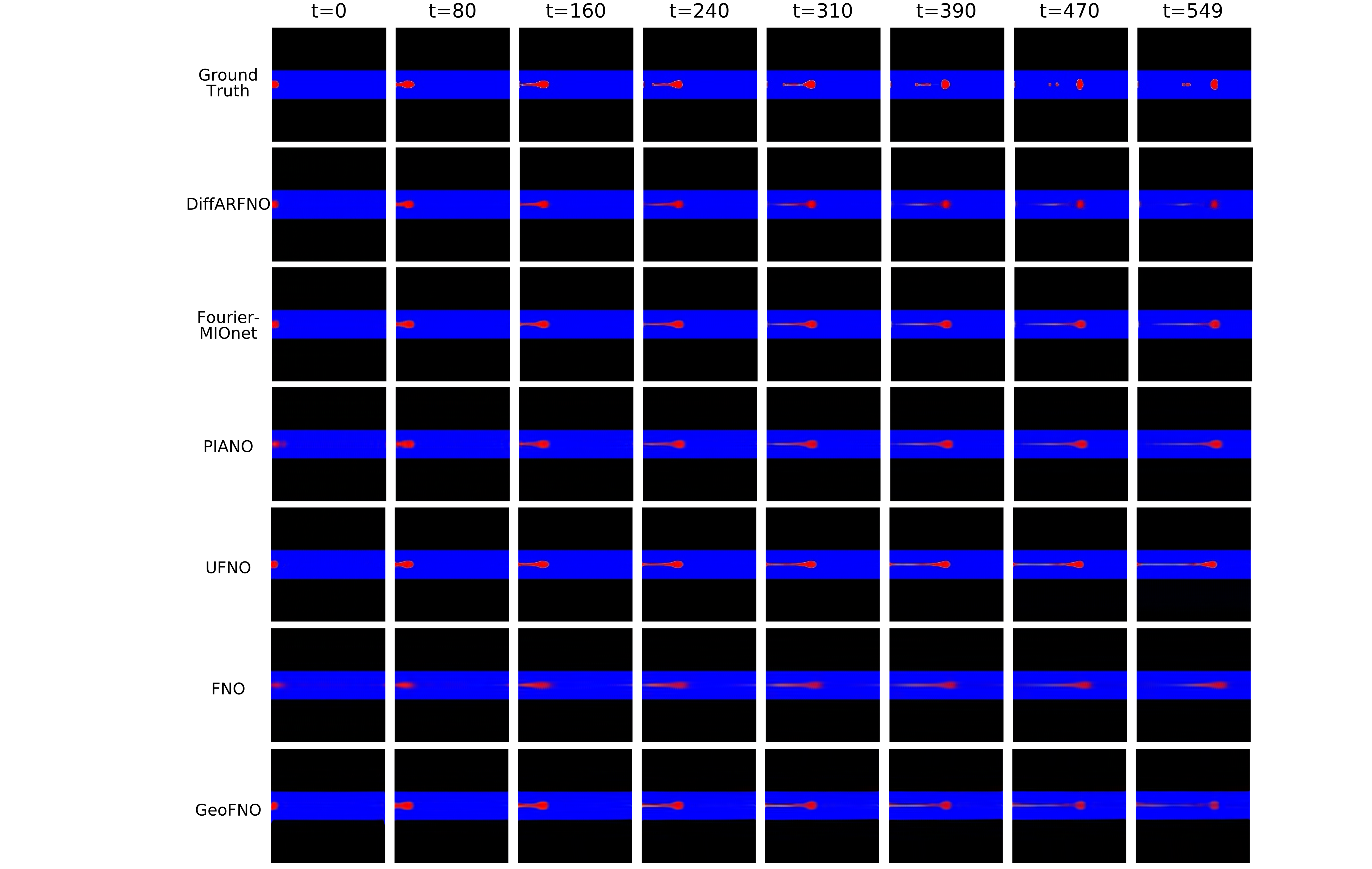}\vspace{-0.05in}
\caption{RGB predictions of \textbf{satellite droplet evolution}.} \vspace{-0.15in}
\label{fig:qual_sample0}
\end{figure}

\subsubsection{\emph{Satellite Droplet Formation}} \label{subsubsec: satellite}

\begin{figure}[!htbp]
\centering
\includegraphics[width=\linewidth]{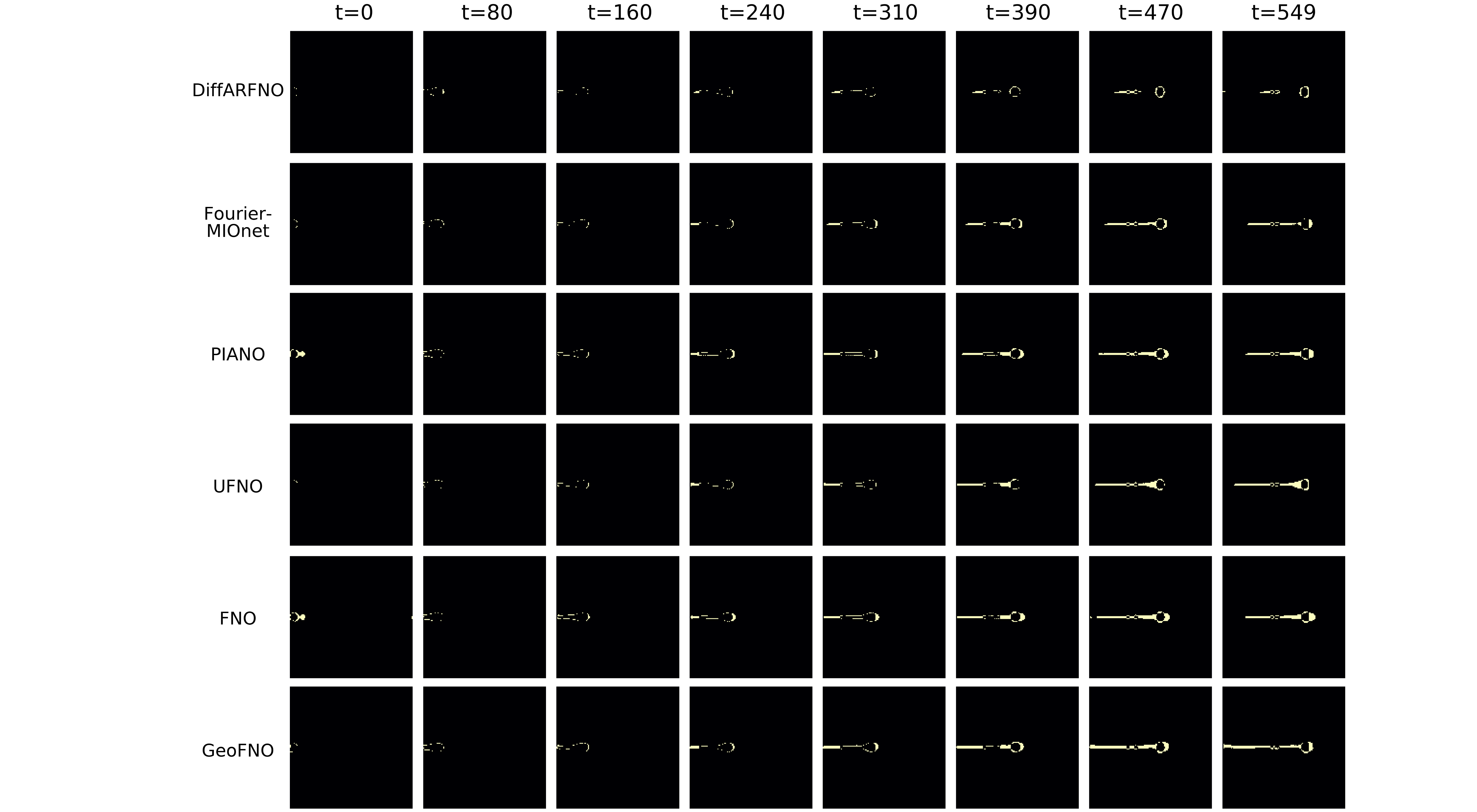}\vspace{-0.05in}
\caption{The error map of binary masks for \textbf{satellite droplet evolution}. The error map is calculated between the prediction mask and the ground truth mask.} \vspace{-0.15in}
\label{fig:qual_sample0_error}
\end{figure}

Figure~\ref{fig:qual_sample0} illustrates a \textbf{satellite droplet formation} regime. The error maps in Figure~\ref{fig:qual_sample0_error} are computed based on the predicted and ground truth masks. This regime is particularly challenging because the breakup from a continuous ligament into multiple small droplets is a transient event; accurately capturing the timing of droplet separation is difficult. In addition, satellite droplets occupy only a small number of pixels. Accurate prediction, therefore, requires the model to capture the transient breakup timing and resolve the emergence of multiple small droplets.

DiffARFNO provides the closest visual agreement with the ground truth in this regime. It better preserves the main droplet trajectory while retaining small detached structures near the expected locations, indicating the sensitivity to fine-scale details. Although some of the small satellite features are still weakened, DiffARFNO captures the satellite formation pattern more faithfully than the benchmark models in both RGB images and binary masks.

In contrast, Fourier-MIONet captures the global droplet motion but tends to smooth out the smallest satellite droplets, leading to missing satellite droplet in the RGB and masks. PIANO produces plausible main-body morphology, but the small separated droplets are weakened by the surrounding ligament structure. UFNO and FNO show more noticeable loss of satellite features, with predictions dominated by a smoothed main ligament and terminal droplet. GeoFNO also struggles in this regime, showing substantial smoothing and limited recovery of the detached small-scale droplets.

\begin{figure}[!htbp]
\centering
\small

\begin{subfigure}[t]{1\linewidth}
    \centering
    \includegraphics[width=\linewidth]{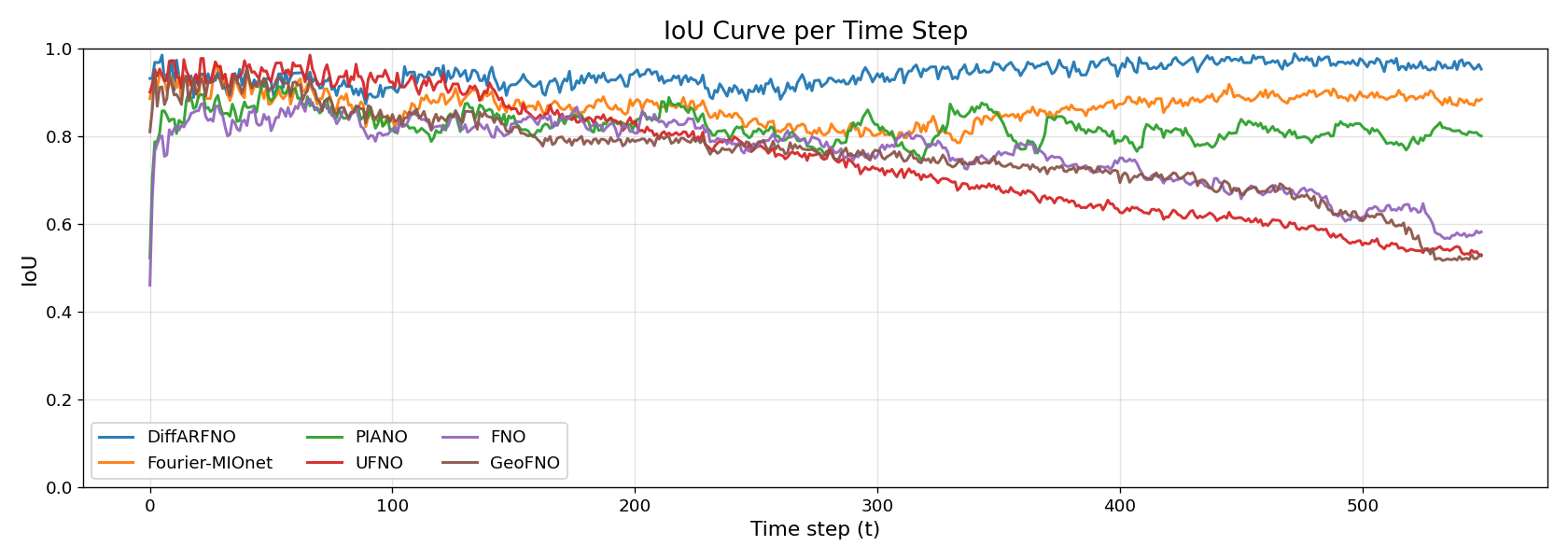}\vspace{-0.1in}
    \caption{Ligament thinning: frame-wise IoU over the prediction horizon.}
    \label{fig:iou_thinneck}
\end{subfigure}

\begin{subfigure}[t]{1\linewidth}
    \centering
    \includegraphics[width=\linewidth]{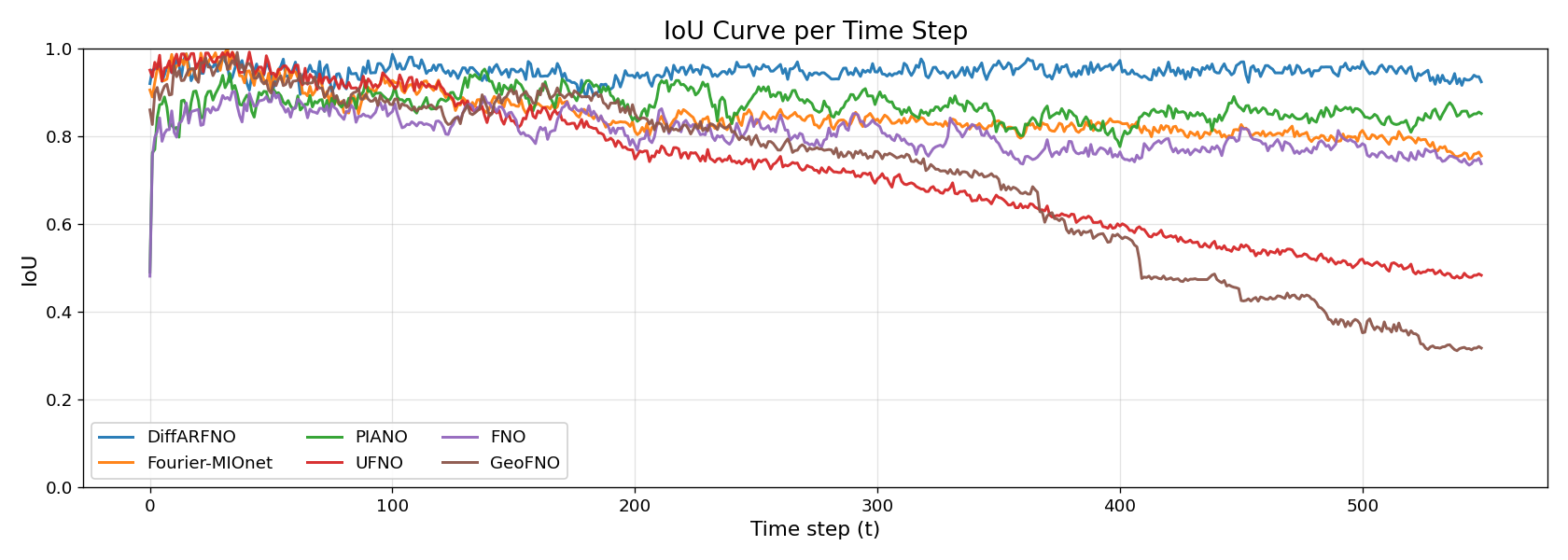}\vspace{-0.1in}
    \caption{Pinch-off dynamics: frame-wise IoU over the prediction horizon.}
    \label{fig:iou_pinchoff}
\end{subfigure}

\begin{subfigure}[t]{1\linewidth}
    \centering
    \includegraphics[width=\linewidth]{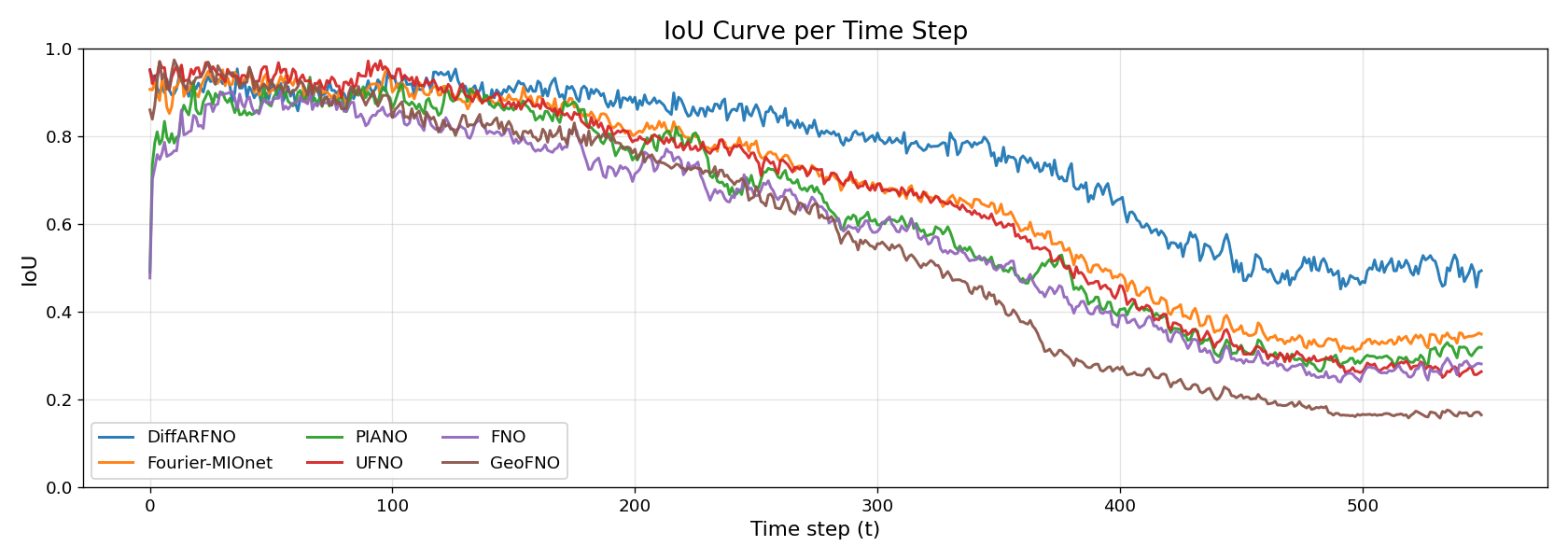}\vspace{-0.1in}
    \caption{Satellite droplet formation: frame-wise IoU over the prediction horizon.}
    \label{fig:iou_satellite}
\end{subfigure}
\vspace{-0.1in}
\caption{Frame-wise IoU trajectories for three representative test samples corresponding to distinct droplet evolution regimes. Each curve reports segmentation overlap at each time step for different benchmark models.}
\label{fig:iou_curves}
\end{figure}

\subsection{\emph{Performance Evaluation on Temporal Evolution}}
\label{sec:mask_iou}

To evaluate segmentation accuracy over time, we analyze frame-wise IoU for the same three representative test samples used in Section \ref{sec:qual_results}. Figure~\ref{fig:iou_curves} summarizes the IoU evolution over the full prediction horizon for DiffARFNO and the benchmark models.

\subsubsection{\emph{Ligament Thinning}}
 Figure~\ref{fig:iou_thinneck} shows that DiffARFNO maintains consistently high IoU values throughout the sequence and shows limited temporal fluctuation.  Fourier-MIONet also remains stable and achieves the strongest baseline performance, although its IoU stays below that of DiffARFNO during most later time steps. PIANO maintains moderate-to-high IoU but exhibits larger fluctuations. In contrast, GeoFNO, UFNO, and FNO show clearer long-horizon deterioration, where the IoU has a decreasing trend along the prediction horizon. 

\subsubsection{\emph{Pinch-Off Dynamics}}

Figure~\ref{fig:iou_pinchoff} shows that DiffARFNO maintains the highest IoU for most of the sequence, demonstrating improved robustness near localized neck thinning and topology-transition regions. FNO, Fourier-MIONet, and PIANO retain relatively stable segmentation accuracy at later time steps, with PIANO slightly outperforming Fourier-MIONet and FNO in parts of the later horizon.  In contrast, UFNO and GeoFNO exhibit substantial degradation over time, where both models show a sharp decline at a late stage, reflecting accumulated spatial mismatch during autoregressive prediction.

\subsubsection{\emph{Satellite Droplet Formation}}

As we mentioned in Section~\ref{subsubsec: satellite}, the satellite droplet formation regime is particularly challenging. Especially, accurately capturing the timing of droplet separation is difficult. Compared with the other two regimes, the IoU curves of all models in Figure~\ref{fig:iou_satellite} decrease more rapidly as small secondary droplets emerge and separate from the main body. DiffARFNO consistently achieves the highest IoU across the sequence and shows a slower degradation rate than the benchmark models, although its performance also declines over time. Fourier-MIONet is the strongest baseline in this regime, while PIANO, FNO, and UFNO exhibit larger drops as the satellite structures become more difficult to localize. GeoFNO performs the worst, with the steepest decline and the lowest late-stage IoU, indicating limited ability to preserve small detached droplets.


\subsection{\emph{Velocity Evaluation for Droplet Evolution}}
\label{sec:headpos_speed_eval}

To assess the physical fidelity of the predicted droplet evolution beyond pixel-level reconstruction accuracy, we perform a velocity-based evaluation by extracting head positions and velocity from both ground truth and predicted droplet evolution.

\subsubsection{\emph{Head Position Extraction and Velocity Estimation}}
\begin{figure}[!htbp]
    \centering
    \includegraphics[width=0.8\linewidth]{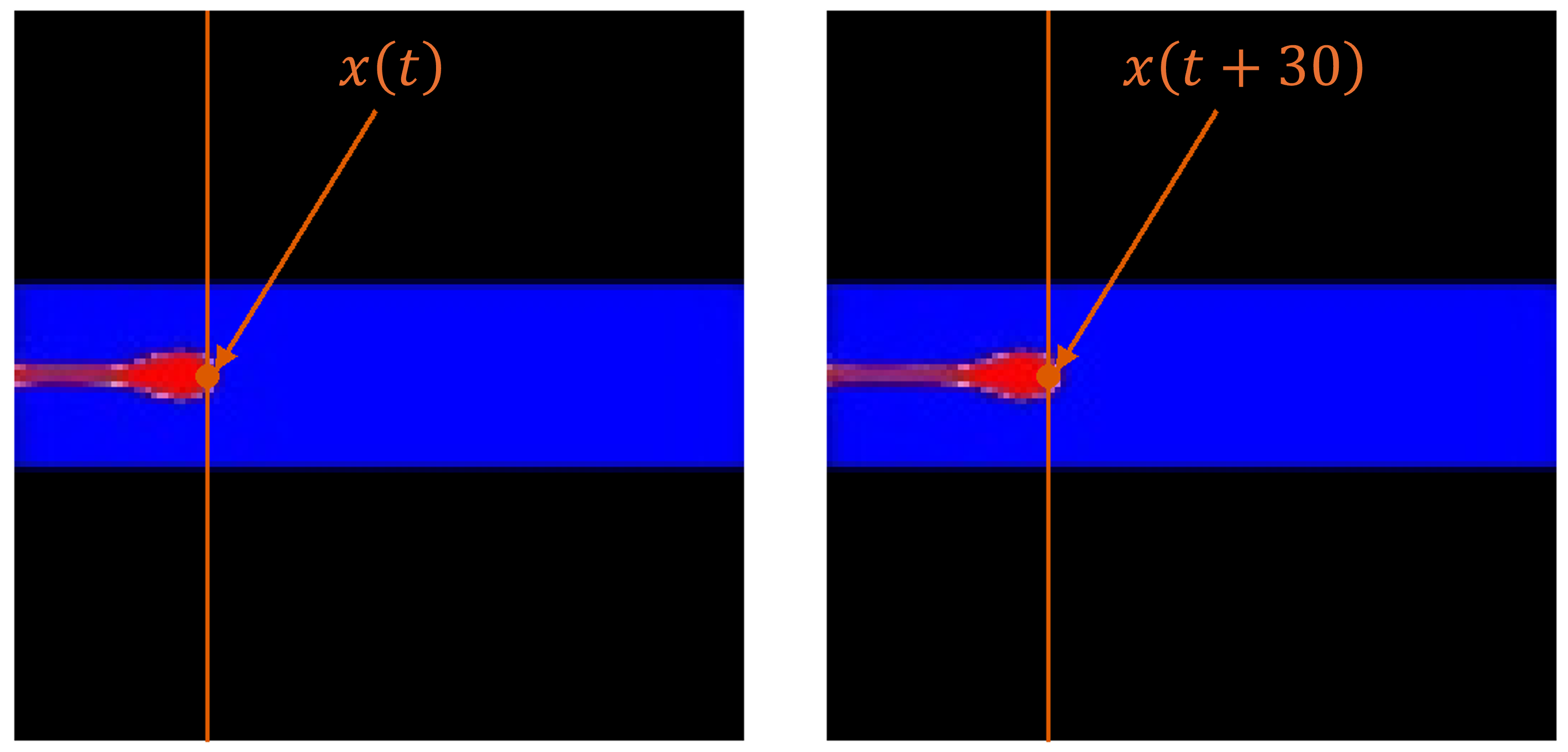}
    \caption{Illustration of droplet head positions used for velocity calculation. The droplet-head position $x(t)$ is extracted from two frames separated by $K=30$ time steps, and the change in position is used to estimate the forward velocity. }
    \label{fig:head_velocity_estimation}
\end{figure}

For the right-moving droplets considered here, we compute the column-wise foreground mass using binary weights. The head position $x(t)$ is defined as the horizontal coordinate at which the cumulative mass reaches a fraction $q$ of the total mass. We use $q=0.995$, so $x(t)$ tracks the forward head location while being less sensitive to isolated boundary pixels. Figure~\ref{fig:head_velocity_estimation} visualizes how the droplet-head position $x(t)$ is extracted for velocity calculation. To reduce frame-to-frame noise, the resulting evolution is smoothed by a one-dimensional Gaussian filter, producing $\tilde{x}(t)$.
We then compute the forward-$K$ velocity on the \emph{start-time} axis:
\begin{equation}
v(t)=\frac{\tilde{x}(t+K)-\tilde{x}(t)}{K\,\Delta t},
\qquad t\in\{0,\ldots,N-K-1\},
\label{eq:forwardK_vx}
\end{equation}
where $K=30$ in the reported evaluation and $\Delta t$ is the time step. The last $K$ frames are invalid because $x(t+K)$ is unavailable. The velocity sequence is smoothed with another one-dimensional Gaussian filter to reduce small velocity fluctuations caused by taking differences between noisy position estimates before computing the reported metrics.

\subsubsection{\emph{Quantitative Evaluation of Droplet Velocity}} 

For each test sample (one video), we first compute the velocity metric using all valid time steps in that sample. The reported value is then obtained by averaging the per-sample metric values across all evaluated samples. Table~\ref{tab:vx_metrics} summarizes the resulting velocity errors of different models, using MAE, Root Mean Squared Error (RMSE), and Relative Error (RE). 

\begin{table}[!htbp]
    \centering
    \caption{Performance evaluation of velocity estimation. Metrics use forward velocity with $K=30$ and are averaged across test samples. }
    \label{tab:vx_metrics}
    \begin{tabular}{l c c c c c}
        \toprule
        Model & MAE & RMSE & RE \\
        \midrule
        DiffARFNO & \textbf{0.0117} & \textbf{0.0202}  & \textbf{0.1190} \\
        Fourier-MIONet & 0.0191 & 0.0249 & 0.1646 \\
        PIANO & 0.0225 & 0.0335 & 0.1731 \\
        UFNO & 0.0267 & 0.0371 & 0.1844  \\
        FNO & 0.0655 & 0.0757 & 0.4747 \\
        GeoFNO & 0.0204 & 0.0297 & 0.1714  \\
        \bottomrule
    \end{tabular}
\end{table}

Compared with the benchmarks, \textbf{DiffARFNO} achieves the most accurate velocity estimation, yielding the lowest MAE of $0.0117$, RMSE of $0.0202$, and RE of $0.1190$, indicating strong agreement with the ground truth droplet velocity. Fourier-MIONet achieves the second-best MAE, RMSE, and RE, suggesting that its overall velocity estimation is quite accurate. Specifically, Fourier-MIONet achieves an MAE of $0.0191$ while DiffARFNO has an MAE of $0.0117$, reducing the MAE by approximately $38.7\%$. Fourier-MIONet has an RMSE of $0.0249$ while DiffARFNO reduces the RMSE by approximately $18.9\%$. Fourier-MIONet has an RE of $0.1646$ while DiffARFNO reduces the RE by approximately $27.7\%$. 

Taken together, these results show that DiffARFNO provides the most reliable velocity estimation among all the benchmark models. This indicates that the proposed framework not only improves pixel-level reconstruction quality but also better preserves the physical information, i.e., the velocity of the droplet.


\section{Conclusion}\label{s:conclusion}

We present \textbf{DiffARFNO}, a two-stage framework for high-fidelity, long-horizon prediction of droplet evolution. Our framework couples an autoregressive Fourier-MIONet, capturing coarse dynamics, and a DDIM corrector that restores fine details. By doing so, our DiffARFNO effectively captures complex topological transitions, such as pinch-off and satellite formation, that are challenging for existing data-driven models. We evaluated our model on a comprehensive dataset generated using ANSYS Fluent. The results show that our DiffARFNO achieves the best performance across both pixel-wise and structural metrics among benchmark models. This work establishes a rapid, high-fidelity model that significantly reduces the computational overhead of droplet simulations. 

Despite these promising results, this study is limited by its reliance on CFD-generated data, and the transferability of the trained model to real experimental additive manufacturing conditions remains to be validated. Future work will focus on integrating this framework into real-time monitoring systems, improving the transferability of data-driven models to real-world additive manufacturing processes. Beyond inkjet printing, extending this framework toward digital twin applications could support process control and optimization in other manufacturing processes \citep{wang2025smartfixture, liu2025quantum, lutz2026reinforcement}.



\bibliographystyle{chicago}
\spacingset{1}
\bibliography{IISE-Trans}

\newpage
\appendix
\setcounter{table}{0}
\section{Training Hyperparameters and Implementation Settings} 
\label{appendix: hyperparameters}

This appendix summarizes the main training hyperparameters and implementation settings used for the proposed framework and all benchmark models. The backbone prediction model and the diffusion correction module are trained separately, with their optimizer, learning rate, scheduler, autoregressive window configuration, and early-stopping settings listed in Table~\ref{tab:train_hparams}.

\begin{table}[!htbp]
\centering
\caption{Training hyperparameters and implementation settings.}
\label{tab:train_hparams}
\small
\begin{tabular}{l l}
\hline
\textbf{Item} & \textbf{Value} \\
\hline
Backbone optimizer & Adam \\
Backbone learning rate & $1\times10^{-4}$ \\
Backbone epochs & 30 \\
Backbone early-stopping patience & 8 \\
LR scheduler & StepLR (step\_size=15, $\gamma$=0.5) \\
\hline
Diffusion optimizer & AdamW \\
Diffusion learning rate & $2\times10^{-4}$ \\
Diffusion epochs & 30 \\
Diffusion early-stopping patience & 8 \\
Diffusion LR scheduler & CosineAnnealingLR ($T_{\max}=30$) \\
Diffusion timesteps $T_d$ & 200 \\
Noise schedule & $\beta_t\in[10^{-4},\,2\times10^{-2}]$ \\
DDIM sampling steps & 25 \\
\hline
First window input (Ground Truth) & 50 frames \\
Sliding window size & 50 frames \\
Per-window output prediction length & 50 frames \\
\hline
\end{tabular}
\end{table}

\subsection{Fourier-MIONet Implementation}

The coarse predictor follows the structure of Fourier-MIONet. The input video sequence is first lifted to the feature space by a point-wise $1\times1\times1$ convolution and then processed by a three-block 3D FNO stack. In each FNO block, a spectral convolution branch and a point-wise convolution branch are summed, followed by a $1\times1\times1$ channel-mixing convolution layer. The result is added back to the block input through a residual connection and normalized by instance normalization. In parallel, the material property vector is embedded by a two-layer MLP, and the normalized coordinates are encoded using Fourier positional features and projected with a point-wise $1\times1\times1$ convolution. The video, property, and coordinate features are then concatenated and passed to a UFNO-style decoder. 

The decoder first maps the concatenated features to the decoder width using a $1\times1\times1$ convolution. It then applies two downsampling stages, a bottleneck FNO stack, and two upsampling stages. Each downsampling, bottleneck, and upsampling stage contains two 3D FNO blocks. The decoded features are projected back to the final prediction using an output head composed of two $1\times1\times1$ convolutions with GELU activation.


\subsection{DDIM Implementation}

The second-stage corrector consists of a conditional 3D U-Net denoiser and a DDIM sampling process. The denoiser uses a 3D U-Net with two downsampling stages and two upsampling stages. The diffusion timestep is encoded by a sinusoidal embedding and injected into the convolutional blocks, which use group normalization and SiLU activation. A final $1\times1\times1$ convolution maps the decoded features to the output noise channels. We use the diffusion process with 200 timesteps during training and use a 25-step DDIM sampler at inference.

\begin{figure}[!htb]
\centering
\includegraphics[width=\linewidth]{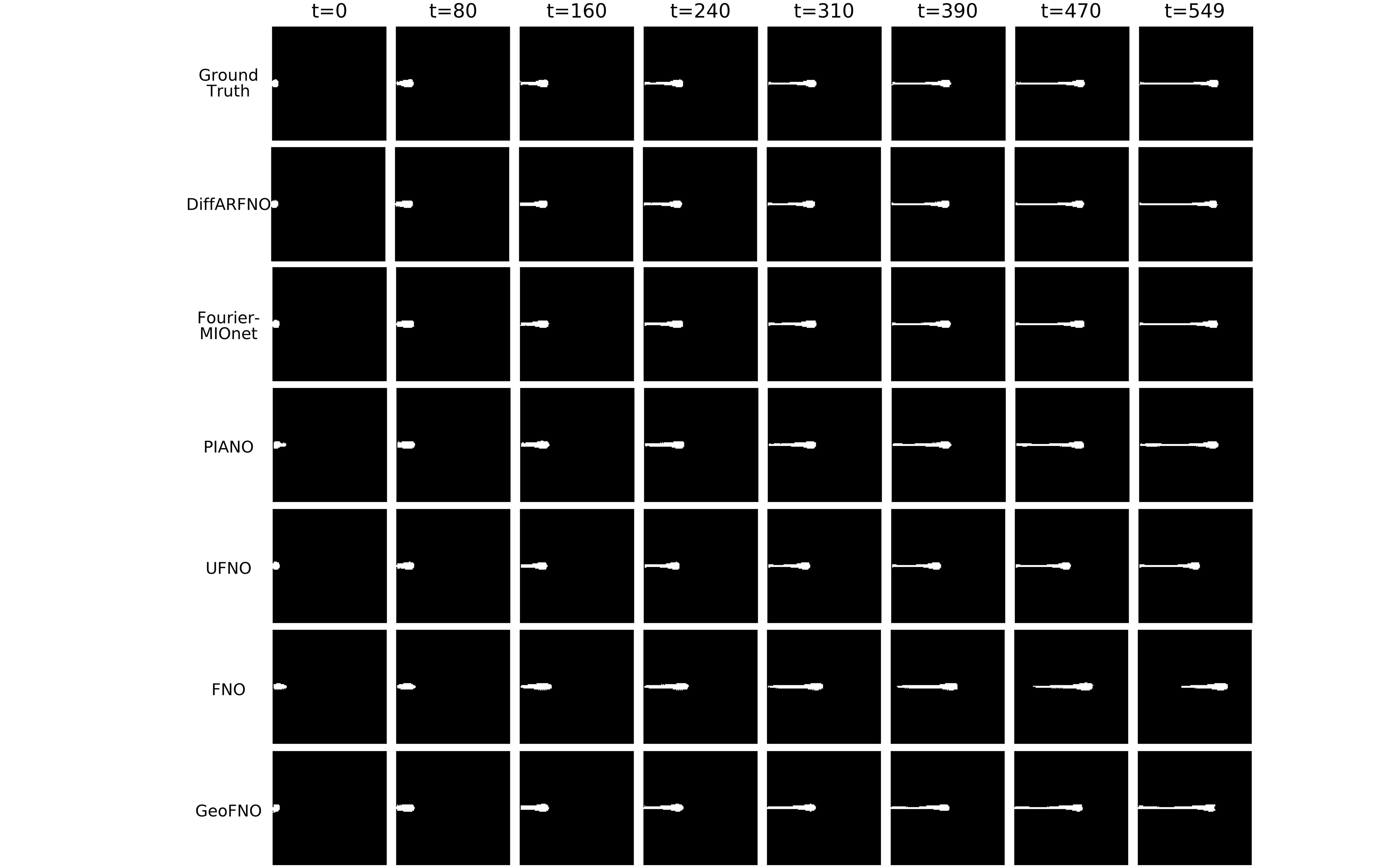} 
\caption{\textbf{Ligament thinning regime.} Binary mask predictions for a representative case with an increasingly thinning ligament. DiffARFNO preserves the ligament thinning structure and boundary more accurately than the compared benchmarks.} 
\label{fig:Sample41Mask}
\end{figure}

\section{Binary Mask Predictions}\label{appendix: binary}

\begin{figure}[!htb]
\centering
\includegraphics[width=\linewidth]{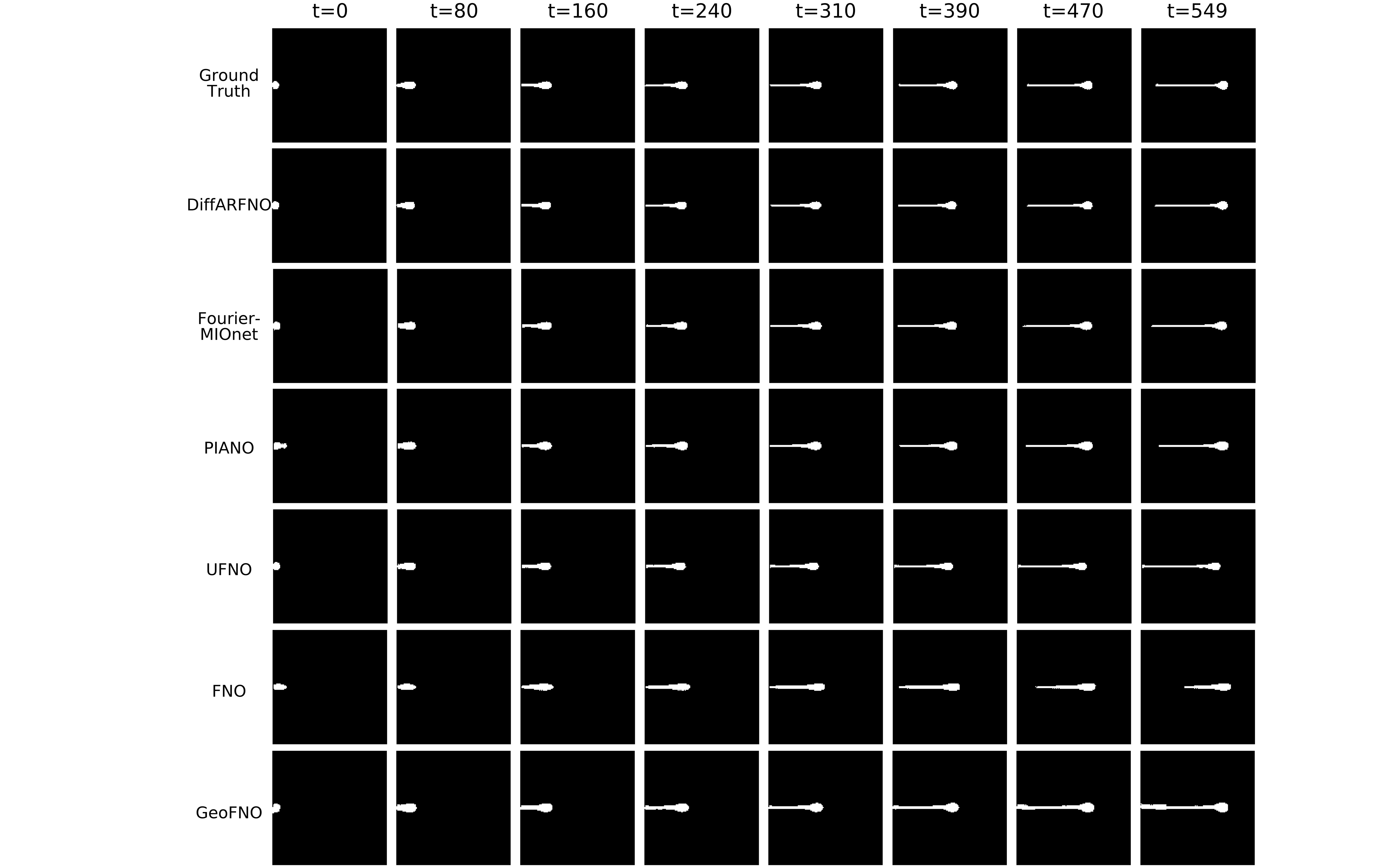} 
\caption{\textbf{Pinch-off dynamics regime.} Binary mask predictions for a representative topology-changing case. DiffARFNO better captures the pinch-off dynamics, whereas the benchmarks show errors such as missed pinch-off and neck morphology errors.}
\label{fig:Sample39Mask}
\end{figure}

\begin{figure}[!htp]
\centering
\includegraphics[width=\linewidth]{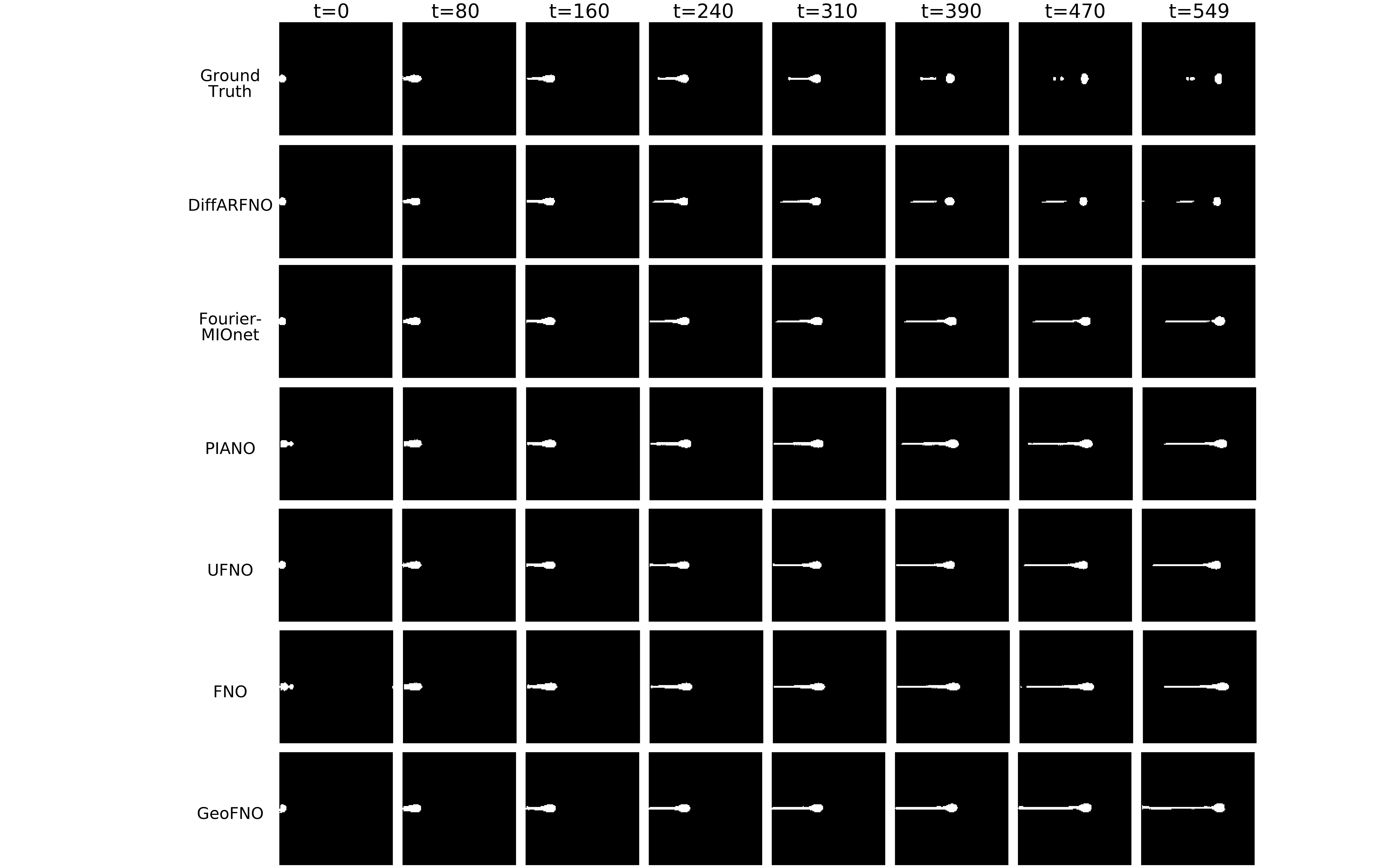}
\caption{\textbf{Satellite droplet formation regime.} Binary mask predictions for a representative case involving small detached droplets. DiffARFNO more accurately recovers the satellite droplets and the main ligament morphology, while the benchmarks fail to detect the small droplets or predict them with inaccurate morphology.} 
\label{fig:Sample0Mask}
\end{figure}

To further show the performance of the proposed model under different regimes, additional binary-mask results are provided for the same three samples discussed in Section~\ref{sec:qual_results}. Figures~\ref{fig:Sample41Mask}--\ref{fig:Sample0Mask} show the corresponding three representative regimes: ligament thinning, pinch-off dynamics, and satellite droplet formation. 

For ligament thinning in Figure~\ref{fig:Sample41Mask}, DiffARFNO most consistently preserves the slender ligament and produces the sharpest boundary among the benchmarks. PIANO and Fourier-MIONet also generate reasonable binary masks; however, both exhibit small but visible errors around the boundary. In contrast, the other benchmarks tend to over-smooth the ligament, shorten the ligament, or introduce local discontinuities near the thin tail. These errors indicate that they struggle to maintain stable fine-scale structures once the droplet becomes slender.

For pinch-off dynamics in Figure~\ref{fig:Sample39Mask}, DiffARFNO better captures the thinning process and the subsequent pinch-off of the droplet from the ligament. The predicted masks remain close to the reference in both the neck width and the morphology of the droplet head. PIANO and Fourier-MIONet also recover the main pinch-off structure, but small errors remain near the thinning neck and detached droplet boundary. In contrast, the other benchmarks show more morphological and topological errors, including failure to capture pinch-off, thickening of the neck region, and over-smoothing of the ligament. These errors indicate that they are less reliable when the droplet undergoes a topology-changing event.

For satellite droplet formation in Figure~\ref{fig:Sample0Mask}, DiffARFNO provides the most reliable reconstruction of the small detached droplets while preserving the main ligament morphology. PIANO captures the main ligament morphology but fails to recover the satellite droplets, whereas Fourier-MIONet detects part of the satellite structure but with substantial boundary and localization errors. The remaining benchmarks fail to capture the satellite droplets. This comparison shows that DiffARFNO is better able to preserve fine-scale droplet structures that are easily missed by the other methods.

Overall, these binary-mask results show that DiffARFNO better preserves both the topology and morphology of the droplet structures across different regimes. The remaining methods can reproduce the coarse droplet shape in some cases, but they show visible errors in boundary thickness and location, ligament continuity, and satellite droplet formation. These results indicate that DiffARFNO is more reliable for long-horizon prediction of fine-scale droplet evolution.

\end{document}